\documentclass{article}

\usepackage[preprint]{neurips_2026}

\usepackage[utf8]{inputenc}
\usepackage[T1]{fontenc}
\usepackage{hyperref}
\usepackage{xcolor}
\definecolor{paperred}{RGB}{105, 20, 45}
\hypersetup{
    colorlinks=true,
    linkcolor=paperred,
    citecolor=paperred,
    urlcolor=paperred,
    linktocpage=true,
    breaklinks=true
}
\usepackage{url}
\usepackage{booktabs}
\usepackage{amsfonts}
\usepackage{nicefrac}
\usepackage{microtype}

\usepackage{enumitem}

\usepackage{caption}

\usepackage{amsmath}
\usepackage{amssymb}
\usepackage{mathtools}
\usepackage{amsthm}
\usepackage{thmtools}
\usepackage{etoolbox}

\usepackage{amsfonts}
\usepackage{mathrsfs}
\usepackage{dsfont}
\usepackage{bbm}

\usepackage{relsize}
\usepackage{nicefrac}
\usepackage{centernot}

\usepackage{accents}

\usepackage{algorithm}

\usepackage{algpseudocode}

\usepackage[capitalise,noabbrev,nameinlink,sort]{cleveref} % Added cleveref -- easy referencing

\theoremstyle{plain}
\newtheorem{theorem}{Theorem}[section]
\newtheorem{lemma}[theorem]{Lemma}

\newtheorem{corollary}[theorem]{Corollary}

\theoremstyle{definition}
\newtheorem{definition}[theorem]{Definition}

\theoremstyle{remark}

\AtBeginEnvironment{proof}{\setcounter{localclaim}{0}}



\newcommand{\widebar}[1]{\accentset{\rule{0.9em}{0.75pt}}{#1}}

\DeclareMathOperator*{\argmin}{arg\,min}

\DeclareMathOperator{\VCD}{VC}
\DeclareMathOperator{\SETVC}{SET-VC}
\DeclareMathOperator{\PERTVC}{PERT-VC}
\DeclareMathOperator{\LD}{LD}
\DeclareMathOperator{\OPT}{opt}
\DeclareMathOperator{\Nat}{\mathbb N}

\newcommand{\KL}{\operatorname{D}_{\operatorname{KL}}}

\newenvironment{proofof}[1]{%
  \par\noindent\textbf{Proof of #1. }%
}{
\qed
}

\usepackage{tcolorbox}

\tcbset{
    rounded corners,
    colback = white,
    before skip = 0.25cm,
    after skip = 0.5cm,
    boxrule = 2pt,
    arc = 10pt
}

\usepackage{tikz}

\newcommand{\circled}[1]{
  \tikz[baseline=(char.base)]{
    \node[
      shape=circle,
      draw,
      inner sep=0.4pt,
      minimum size=1.2em
    ] (char) {\strut #1};
  }
}

\newcommand{\customcal}[1]{\mathcal{#1}}
\newcommand{\cA}{\customcal{A}}
\newcommand{\cB}{\customcal{B}}
\newcommand{\cC}{\customcal{C}}
\newcommand{\cD}{\customcal{D}}

\newcommand{\cL}{\customcal{L}}

\newcommand{\cQ}{\customcal{Q}} 
 
\newcommand{\cS}{\customcal{S}}

\newcommand{\cW}{\customcal{W}}
\newcommand{\cX}{\customcal{X}}
\newcommand{\cY}{\customcal{Y}}

\newcommand{\bA}{\mathbf{A}}

\newcommand{\bR}{\mathbf{R}}
\newcommand{\bM}{\mathbf{M}}

\crefname{section}{Section}{Sections}
\crefname{theorem}{Theorem}{Theorems}
\crefname{lemma}{Lemma}{Lemmas}
\crefname{problem}{Problem}{Problems}
\crefname{program}{Program}{Progams}
\crefname{definition}{Definition}{Definitions}
\crefname{conjecture}{Conjecture}{Conjectures}
\crefname{corollary}{Corollary}{Corollaries}
\crefname{construction}{Construction}{Constructions}
\crefname{conjecture}{Conjecture}{Conjectures}
\crefname{claim}{Claim}{Claims}
\crefname{observation}{Observation}{Observations}
\crefname{proposition}{Proposition}{Propositions}
\crefname{fact}{Fact}{Facts}
\crefname{question}{Question}{Questions}
\crefname{problem}{Problem}{Problems}
\crefname{remark}{Remark}{Remarks}
\crefname{example}{Example}{Examples}
\crefname{equation}{Equation}{Equations}
\crefname{appendix}{Section}{Sections}
\crefname{algorithm}{Algorithm}{Algorithms}
\crefname{model}{Model}{Models}
\crefname{figure}{Figure}{Figures}
\crefname{infassumption}{Informal Assumption}{Informal Assumptions}
\crefname{inftheorem}{Informal Theorem}{Informal Theorems}
\crefname{infdefinition}{Informal Definition}{Informal Definitions}
\crefname{minftheorem}{Main Informal Theorem}{Main Informal Theorems}
\crefname{maintheorem}{Main Theorem}{Main Theorems}
\crefname{assumption}{Assumption}{Assumptions}
\crefname{step}{Step}{Steps}
\crefname{result}{Result}{Results}
\crefname{event}{Event}{Events}
\crefname{none}{}{}

\newcommand{\inbrace}[1]{\left\{#1\right\}}
\newcommand{\inparen}[1]{\left(#1\right)}
\newcommand{\insquare}[1]{\left[#1\right]}

\newcommand{\floor}[1]{\left\lfloor#1\right\rfloor}

\newcommand{\prob}[1]{\Pr\insquare{#1}}

\newcommand{\expect}[1]{\mathbb{E} \insquare{#1}}
\newcommand{\cexpect}[2]{\mathbb{E}_{#1}\insquare{#2}}

\usepackage[suppress]{color-edits}
\addauthor[Amirreza]{as}{gold}
\addauthor[Farnam]{fm}{red}
\addauthor[Shai]{sbd}{blue}
\addauthor[Steve]{sh}{cyan}

\title{Near-Optimal Sample Complexity of Deterministic Agnostic PAC Learning}

\author{
    Shai Ben-David\\
    David Cheriton School of Computer Science\\
    University of Waterloo \& Vector Institute\\
    Waterloo, ON, Canada\\
    \texttt{shai@uwaterloo.ca}
    \And
    Steve Hanneke\\
    Department of Computer Science\\
    Purdue University\\
    West Lafayette, IN, USA\\
    \texttt{steve.hanneke@gmail.com}\\
    \And
    Farnam Mansouri\\
    David Cheriton School of Computer Science\\
    University of Waterloo \& Vector Institute\\
    Waterloo, ON, Canada\\
    f5mansou@uwaterloo.ca\\ 
    \And
    Amirreza Shaeiri\\
    Department of Computer Science\\
    Purdue University\\
    West Lafayette, IN, USA\\
    \texttt{amirreza.shaeiri@gmail.com}\\ 
}

\begin{document}

\maketitle

\begin{abstract}
We study the problem of agnostic PAC learning under deterministic labels, where the examples are labeled by an arbitrary deterministic function, which need not belong to the concept class $\mathcal{C}$. This model captures misspecification without stochastic label noise and lies between the realizable and fully agnostic settings. Prior work showed that, when $\mathcal{C}$ has finite diameter $K$ and VC dimension $d$, deterministic labels can improve the sample complexity upper bound from the classical agnostic guarantee $\mathcal{O}(d/\varepsilon^2)$ to $\widetilde{\mathcal{O}}(K/\varepsilon)$, but left a large gap between upper and lower bounds.
Our main result sharpens this picture. We prove that every class of VC dimension $d$ and finite diameter $K$ is learnable with sample complexity
\(
\widetilde{\mathcal{O}}\left(
\frac{K^{1/3} \, d^{2/3}}{\varepsilon}
\right),
\)
and we construct classes for which this bound is tight up to logarithmic factors. 
We also study the online analogue of this problem and show that the minimax regret is again partly controlled by the diameter of the class. Technically, our upper bounds are obtained via memorization-based learning algorithms that exploit the deterministic label structure.
\end{abstract}

\section{Introduction} \label{sec: introduction}

The seminal works of \cite{vapnik1974theory, valiant1984theory} laid the foundations of modern statistical learning theory by introducing the \emph{probably approximately correct (PAC) learning framework}, a fundamental theoretical criterion for supervised learning that has been the focus of extensive research over the past several decades.

In this framework, there is an unknown data distribution $\mathcal{D}$ over $\mathcal{X} \times \mathcal{Y}$, where $\mathcal{X}$ is an instance space and $\mathcal{Y}$ is a binary label space.~\footnote{For clarity of exposition, we omit standard set and measure theoretic qualifications in the introduction. The formal definitions are given in \autoref{sec: definitions}.} For example, $\mathcal{X}$ may be a space of medical images, and $\mathcal{Y} = \{0, 1\}$ may be the binary label space, where $y=1$ indicates that the corresponding image shows a certain abnormality and $y=0$ indicates that it does not. In addition, this framework fixes a concept class $\mathcal{C}$, consisting of functions from the instance space $\mathcal{X}$ to the label space $\mathcal{Y}$, which serves as a benchmark family of classifiers. For instance, $\mathcal{C}$ may be the class of neural networks with a prescribed architecture. Moreover, given a set of i.i.d.\ training examples drawn from $\mathcal{D}$, the learner’s goal is to output a hypothesis $\hat{h} : \mathcal{X} \to \mathcal{Y}$ whose classification error on future examples also drawn from $\mathcal{D}$ is, with high probability, close to that of the best concept in $\mathcal{C}$.

The statistical and computational learning theory communities have studied several assumptions on the relationship between instances and labels, especially within the PAC learning framework. In the \emph{realizable} setting, it is assumed that the labels are generated by a fixed concept $c^\star \in \mathcal{C}$. In the \emph{agnostic} setting, in contrast, no such assumption is imposed: the joint distribution over instances and labels may be arbitrary. Between these two regimes lie \emph{noise models}, such as Tsybakov noise \cite{10.1214/aos/1017939240, 10.1214/aos/1079120131} and Massart noise \cite{10.1214/009053606000000786}, which assume that the labeling rule belongs to $\mathcal{C}$ but allow stochastic label noise of restricted severity. Another natural intermediate assumption is that the \emph{Bayes-optimal classifier} belongs to $\mathcal{C}$. This retains the assumption that the optimal decision rule is contained in the benchmark class but relaxes the control over the label noise.

In this work, we assume that the labels are \emph{deterministic}. That is, there exists a fixed function $f^\star : \mathcal{X} \to \mathcal{Y}$ specifying the label of each instance; however, $f^\star$ need not belong to the concept class $\mathcal{C}$. This model was introduced and systematically studied in the pioneering work of \cite{ben2014sample}. A similar assumption has also been investigated in other statistical learning frameworks \cite{ben2011learning, urner2013probabilistic, 10.1093/imaiai/iaab010}. Moreover, this formulation also studied in the literature on computational learning theory, particularly in the agnostic learning of Boolean concept classes; see, for example, \cite{kearns1992toward, feldman2009agnostic}. Finally, it is closely related to the comparative PAC learning framework, introduced by \cite{hu2022comparative} and subsequently developed in \cite{ashkezari2025refining}, in which the target labeling function $f^\star$ is restricted to belong to a fixed class $\mathcal{H}$.

This assumption is motivated by modern high-dimensional learning tasks, in which instances often contain sufficient information to determine the correct label. For example, in medical imaging, a sufficiently detailed scan, together with relevant patient information, may determine whether a specific abnormality is present. However, the resulting labeling rule can be highly complex and does not need to belong to the chosen concept class $\mathcal{C}$. Thus, this assumption allows the task to have a well-defined ground truth while still permitting model misspecification.

This leads to the following fundamental and natural question:

\begin{center}
    \textit{Given a concept class $\mathcal{C} \subseteq \{0,1\}^{\mathcal{X}}$, how many training samples are necessary and sufficient for agnostically PAC learning $\mathcal{C}$ under deterministic labels?}
\end{center}

The work of \cite{ben2014sample} made significant progress on the above question. In particular, they proved that if the \emph{diameter} of a concept class $\mathcal{C}$, denoted by $\operatorname{D}(\mathcal{C})$, is infinite, then the optimal sample complexity of agnostically PAC learning $\mathcal{C}$ under deterministic labels admits a lower bound of $\Omega \big( \log(1/\delta)/\varepsilon^2 \big)$. Moreover, they showed that, for every $d \in \mathbb{N}$, there exists a concept class $\mathcal{C}$ whose Vapnik--Chervonenkis dimension, denoted by $\operatorname{VC}(\mathcal{C})$, equals $d$, for which the optimal sample complexity of agnostic PAC learning under deterministic labels is lower bounded by $\Omega \left( d/\varepsilon^2 + \log(1/\delta)/\varepsilon^2 \right)$. Indeed, this lower bound matches, up to constant factors, the upper bound for the fully agnostic setup achieved by empirical risk minimizers (ERMs) proved via the chaining technique \cite{10.1214/aop/1176988847}. Thus, the \emph{infinite-diameter regime} already exhibits essentially classical agnostic behavior, leaving limited room for improvement from the deterministic label assumption. In contrast, when $\operatorname{D}(\mathcal{C})$ is finite, they proved an upper bound of $\widetilde{\mathcal{O}}(\operatorname{D}(\mathcal{C})/\varepsilon)$, while the only known lower bound is the standard realizable PAC learning lower bound \cite{ehrenfeucht1989general}, namely $\Omega \big( \operatorname{VC}(\mathcal{C})/\varepsilon + \log(1/\delta)/\varepsilon \big)$ for non-trivial concept classes $\mathcal{C}$. Therefore, closing the gap between the upper and lower bounds in the \emph{finite-diameter regime} remains open.

\subsection{Overview of the Main Results} \label{sec: overview}

In the present work, our main contribution is to improve both the upper and lower bounds in the finite-diameter regime. Formally, we establish the following theorem.

\begin{theorem} \label{theorem: introduction_1}
Let $\mathcal{C} \subseteq \{0, 1\}^{\mathcal{X}}$ be a non-degenerate concept class with $d := \operatorname{VC}(\mathcal{C}) < \infty$ and $K := \operatorname{D}(\mathcal{C}) < \infty$. Then, the following holds.
\begin{equation*}
\begin{aligned}
    \mathbf{m}_{\mathcal{C}}(\varepsilon, \delta)
    & \in \Omega \left( \frac{K^{1/3} \, \bigl( \log(1/\delta) \wedge K \bigr)^{2/3} + d + \log(1/\delta)}{\varepsilon} \right),\\
    \mathbf{m}_{\mathcal{C}}(\varepsilon, \delta)
    & \in
    \widetilde{\mathcal{O}} \left( \frac{K^{1/3} \, d^{2/3} + K^{1/3} \, \bigl( \log(1/\delta) \wedge K \bigr)^{2/3} + \log(1/\delta)}{\varepsilon} \right).
\end{aligned}
\end{equation*}

Moreover, for some finite $K_0$, for every $d \in \mathbb{N}$ and every $K \geq \max\{K_0, d\}$, there exists a concept class $\mathcal{C} \subseteq \{0, 1\}^{\mathcal{X}}$ with $\operatorname{VC}(\mathcal{C}) = d$ and $K = \operatorname{D}(\mathcal{C})$ for which the above upper bound is tight. Namely:
\begin{equation*}
    \mathbf{m}_{\mathcal{C}}(\varepsilon, \delta) \in \Omega \left( \dfrac{K^{1/3} \, d^{2/3} + K^{1/3} \, \bigl( \log(1/\delta) \wedge K \bigr)^{2/3} + \log(1/\delta)}{\varepsilon} \right).
\end{equation*}
Here, $\mathbf{m}_{\mathcal{C}}$ denotes the optimal sample complexity of agnostically PAC learning $\mathcal{C}$ under deterministic labels; see \autoref{def: pac-learnability} for the formal definition. In addition, we use $\widetilde{\mathcal{O}}(.)$ to hide poly-logarithmic factors. Also, for any $a,b \in \mathbb{R}$, we use the notation $a \wedge b := \min\{a, b\}$.
\end{theorem}

\textbf{Technical overview of the upper bound.} Our upper bound is obtained by a learning algorithm that combines an empirical risk minimizer with \emph{memorization}. This contrasts with the algorithm of \cite{ben2014sample}, which is purely based on memorization. Furthermore, our analysis of this algorithm departs significantly from known analyses of PAC learning algorithms and may be of independent interest. In particular, we begin by studying the special case in which the concept class consists of only \emph{two functions}; a key ingredient in this analysis is the concentration inequalities for the missing mass of discrete distributions \cite{mcallester2003concentration, 10.1214/ECP.v18-2359}. We then extend the argument to \emph{finite concept classes} by drawing on ideas from density estimation \cite{devroye2001combinatorial}. Finally, we pass to \emph{general VC classes} by constructing a suitable finite cover using a technique similar to \cite{hopkins2022realizable}. 
% \fmcomment{I think we should change citation} 
For additional details, see \autoref{sec: two-concepts} and \autoref{sec: upper}.

\textbf{Technical overview of the lower bound.} The lower bound is technically more involved. The reason is that we must go beyond the standard realizable PAC learning lower bound; however, under deterministic labels, memorization eliminates uncertainty on every \emph{sampled point}, so the hardness must lie in the \emph{unsampled} part of the support. We place a small total mass uniformly across a large set of points and assign their deterministic labels according to slightly biased coins, with the direction of the bias chosen uniformly at random. In this way, the best concept in the class, which is either all ones or all zeros on these points, has a small advantage in this rare region. We show that, below the stated sample size, detecting the direction of this advantage is information-theoretically impossible. On the other hand, any learner achieving a small excess error would yield a successful tester by taking a majority vote over its predictions on the unseen rare points, giving the desired contradiction. Finally, an application of the probabilistic method fixes one deterministic labeling for which the learner fails with the required probability. For additional details, including a more detailed technical overview, see \autoref{sec: pac-lower}.

This technical use of memorization in our results is connected to a central phenomenon in machine learning: training-data memorization is pervasive. This is explicit in classical methods such as nearest-neighbor classifiers, whose hypotheses essentially store the training examples, and in support vector machines, whose predictors depend on selected training points. It also arises in ensemble methods such as AdaBoost and random forests, which can fit the training data perfectly while achieving strong generalization performance \cite{wyner2017explaining}. In modern overparameterized models, memorization is even more pronounced: deep networks can fit the training data with near-perfect accuracy, even when the labels are completely random \cite{zhang2016understanding}. More recently, memorization has also been empirically demonstrated in large language models and diffusion models, where portions of the training data can sometimes be extracted from the trained model \cite{carlini2021extracting, carlini2023extracting, nasr2025scalable}.

The above phenomena are somewhat counterintuitive from the classical point of view, where learning is often associated with extracting generalizable patterns rather than retaining individual training examples \cite{bishop2006pattern}. Yet, the prevalence of memorization suggests that it is not simply opposed to generalization, but rather is probably a fundamental aspect of learning whose role is still not fully understood theoretically. In this direction, a recent line of work has made significant progress in identifying setups where memorization is necessary for accurate learning by focusing primarily on long-tailed or otherwise structured distributional settings \cite{feldman2020neural, 10.1145/3357713.3384290, 10.1145/3406325.3451131, feldman2025trade}. In contrast, our results shed light on this phenomenon from a complementary, \emph{distribution-free} perspective by characterizing the optimal sample complexity of agnostic PAC learning under deterministic labels. Relatedly, a use of memorization also appears in the literature on optimistically universal learning; see, for instance, \cite{blanchard2022contextual} and the references therein. Nevertheless, this line of work studies the universal learning framework, whereas the present work focuses on \emph{uniform guarantees}.

% Indeed, in the classical PAC learning framework, empirical risk minimizers obtain near-optimal sample complexity guarantees in both the realizable and agnostic settings. Surprisingly, this is no longer true for agnostic PAC learning under deterministic labels. The work of \cite{ben2014sample} exhibited concept classes for which ERM, and more generally any proper learner, is suboptimal. We further investigate this phenomenon by introducing a new combinatorial complexity parameter, teremed the paired-$\operatorname{VC}$ dimension. In particular, we prove a lower bound on the optimal sample complexity of agnostic PAC learning under deterministic labels by proper learners, while leaving the corresponding upper bound as an open problem. This result is presented in \autoref{sec: proper}. Notably, similar distinctions have been observed in other learning-theoretic settings, including partial concept classes, multiclass learning, and adversarially robust learning \cite{daniely2011multiclass, daniely2015multiclass, montasser2019vc, alon2022theory}.

Our next contribution concerns online binary classification within the Littlestone framework \cite{littlestone1988learning}. In particular, we consider the problem of agnostic online learning under deterministic labels. Interestingly, the diameter plays an important role here as well. Formally, we establish the following theorem.

\begin{theorem} \label{theorem: introduction_2}
Let $\mathcal{C} \subseteq \{0, 1\}^{\mathcal{X}}$ be a concept class with
$K := \operatorname{D}(\mathcal{C}) < \infty$. Then, we have:
\begin{equation*}
    \mathbf{R}^\star(\mathcal{C}, \operatorname{T}) \in \mathcal{O}(K).
\end{equation*}
Moreover, for every $K \in \mathbb{N}$, there exists a concept class $\mathcal{C} \subseteq \{0, 1\}^{\mathcal{X}}$ with $\operatorname{D}(\mathcal{C}) = K$ and $\LD(\mathcal{C}) = 1$ for which we have:
\begin{equation*}
    \mathbf{R}^\star(\mathcal{C}, \operatorname{T}) \in \Omega(K).
\end{equation*}
Here, $\mathbf{R}^\star(\mathcal{C}, \operatorname{T})$ denotes the minimax expected
regret as a function of the concept class $\mathcal{C}$ and the time horizon $\operatorname{T}$, and $\LD(\mathcal{C})$ denotes the Littlestone dimension of $\mathcal{C}$; see \autoref{sec: online-framework} for formal definitions.
\end{theorem}

Notably, as in our PAC learning algorithm, \emph{memorization} is also a key ingredient in the online learning algorithm that obtains the above upper bound. For additional details and further results, see \autoref{sec: online-results}.

As mentioned earlier, \cite{ben2014sample} provided a lower bound on the optimal sample complexity of agnostic PAC learning with deterministic labels in the finite-diameter regime. As our final contribution, we express their result in terms of a combinatorial complexity parameter that we denote by $\SETVC$ (see \autoref{def:setVC}). We then complement this lower bound by introducing a larger combinatorial complexity parameter, denoted by $\PERTVC$ (see \autoref{def:pertVC}), which controls the leading $1/\varepsilon^2$-term in our upper bound in the infinite-diameter regime.

\begin{theorem} \label{theorem: introduction_3}
Let $\mathcal{C} \subseteq \{0,1\}^{\mathcal{X}}$ be a concept class such that
\(
\SETVC(\mathcal{C}) \leq \PERTVC(\mathcal{C}) < \infty.
\)
Then, there exists a constant $k^\star \in \Nat \cup \{0\}$, possibly depending on $\mathcal{C}$, for which the following holds.
\[
\mathbf{m}_{\mathcal{C}}(\varepsilon,\delta)
=
\widetilde{\mathcal{O}}\!\left(
    \frac{\PERTVC(\mathcal{C})}{\varepsilon^2}
    +
    \frac{k^\star}{\varepsilon}
\right).
\]
Moreover, the lower bound of \citet{ben2014sample} implies
\[
\mathbf{m}_{\mathcal{C}}(\varepsilon,\delta)
=
\widetilde{\Omega}\!\left(
    \frac{\SETVC(\mathcal{C})}{\varepsilon^2}
\right).
\]
\end{theorem}

We then raise the natural open question of whether the above lower and upper bounds match, in small $\varepsilon$ regime. Formally, we have the following open question.

\begin{restatable}{openquestion}{SetPertb}
Is there a concept class $\mathcal{C}$ such that $\SETVC(\mathcal{C}) \neq \PERTVC(\mathcal{C})$?
\end{restatable}

% \sbdcomment{I think that these parameters should be defined in the main body of the paper and that when stating the Theorem we should ref to the definitions.}

For additional details and 
% further 
results, we refer to
\autoref{sec: infinite-diameter} and to the concluding discussion there.

% The comparison between the PAC and the online results highlights an interesting distinction. Online-to-batch conversions allow one to translate online guarantees into PAC guarantees with comparable dependence on the relevant combinatorial parameters. However, such conversions applied to our online bound would lead to a PAC guarantee depending linearly on the diameter, whereas our direct PAC analysis achieves a much better dependence.

\subsection{Organization} \label{sec: organization}

% \sbdcomment{When discussing the "remainder of the paper" we must describe also the remainder of the organization of the main body of the paper! Also, there must be a statement describing the division of the content between the main body and the appendices.} 

The remainder of the paper is organized as follows. In \autoref{sec: definitions}, we introduce the formal notation and definitions used throughout the paper. Then, as a warm-up, in \autoref{sec: two-concepts}, we prove our upper bounds for concept classes with two concepts. Finally, \autoref{sec: conclusion} concludes the main body.

In the appendix, \autoref{sec: related-work} gives a more extensive discussion of related work. In \autoref{sec: upper} and \autoref{sec: pac-lower}, we give our remaining PAC learning upper and lower bounds, respectively, for the finite-diameter regime. Then, in \autoref{sec: infinite-diameter}, we present our results for the infinite-diameter regime. Eventually, in \autoref{sec: deferred-definitions} and \autoref{sec: online-results}, we give the definitions and the results for online learning.

\section{Notation and Definitions} \label{sec: definitions}

In this section, we first set our basic notation in \autoref{sec: notation}. We then define the PAC learning framework in \autoref{sec: pac-framework}. Finally, we give the definitions of the main combinatorial complexity parameters considered in this work in \autoref{sec: pac-parameters}.

\subsection{Notation} \label{sec: notation}

In this subsection, we present the basic notations used throughout the paper, most of which are standard in the community. Let $\mathbb{N}$ and $\mathbb{R}$ stand for the set of natural numbers and real numbers, respectively. In addition, for a given $n \in \mathbb{N}$, we use $[n]$ to denote $\big \{ 1, 2, \ldots, n \big \}$. For any $a,b \in \mathbb{R}$, we use the notation $a \wedge b := \min\{a, b\}$. Also, for $a,b \in \mathbb{N}$, we write $a \mid b$ if $a$ divides $b$, that is, if there exists $m \in \mathbb{N}$ such that $b = am$. Next, given $n \in \mathbb{N}$, for any sequence of size $n$ or $n$-tuple $v$, and any $i \in [n]$, let us use $v_i$ to denote the $i$-th element in $v$. Afterward, we denote by $A \times B$ the Cartesian product of two arbitrary sets $A$ and $B$. In addition, for any set $A$ and any $n \in \mathrm{N}$, we let $A^n$ indicate $n$ times the Cartesian product of $A$ with itself. Note that for any set $A$, we define $A^0 := \{\emptyset\}$. Also, given a set $A$, we denote by $A^{*}$ the set of all finite sequences of members of $A$; more formally, $A^{*} := \bigcup^{\infty}_{T = 0} A^T$. Then, for arbitrary sets $X$ and $Y$, we use $Y^X$ to denote the space of all functions from $X$ to $Y$. Going further, let $\mathrm{e}$ and $\log{(\cdot)}$ stand for Euler's number and the Logarithm function in the base $2$, respectively. Finally, we use $\mathcal{O}(\cdot)$, $\Omega(\cdot)$, and $\Theta(\cdot)$ to denote the standard asymptotic notation in theoretical computer science. Furthermore, we use $\widetilde{O}(\cdot)$, $\widetilde{\Omega}(\cdot)$, and $\widetilde{\Theta}(\cdot)$ to hide poly-logarithmic factors.

\subsection{PAC Learning Framework} \label{sec: pac-framework}

We begin by giving the problem setup. Fix a non-empty measurable space $(\mathcal{X}, \Sigma_{\mathcal{X}})$, referred to as the instance space. For example, the instance space $\mathcal{X}$ can be a Euclidean space. In addition, fix $\mathcal{Y} = \{0, 1\}$ as the label space. A concept is a measurable function from $\mathcal{X}$ to $\mathcal{Y}$. With this in mind, fix a non-empty collection of such functions $\mathcal{C}$ as the concept class. For simplicity, we specify instances of the PAC learning framework solely based on the concept class $\mathcal{C}$, which is defined relative to the measurable structure $(\mathcal{X}, \Sigma_{\mathcal{X}})$. 

Additionally, we say that a concept class $\mathcal{C} \subseteq \mathcal{Y}^{\mathcal{X}}$ is non-degenerate if there exist two concepts $h^0, h^1 \in \mathcal{C}$ and distinct points $x_0, x_1, \ldots, x_{k_0} \in \mathcal{X}$ such that $h^0(x_0) = h^1(x_0)$, and $h^0(x_i) \neq h^1(x_i)$ for every $i \in [k_0]$, where $k_0 \in \mathbb{N}$ is constant that appears in \autoref{sec: lower-proof}.

Next, we present the definition of a data distribution. A data distribution $\mathcal{D}$ is a probability measure over $\mathcal{X} \times \mathcal{Y}$. For example, it may be a probability measure over the space of words in a dictionary and their corresponding binary labels.

Subsequently, we present the definition of a learning algorithm. A PAC learning algorithm $\mathbf{A}$ is a mapping from $(\mathcal{X} \times \mathcal{Y})^{*} \times \mathcal{X}$ to $\mathcal{Y}$ whose restriction to $(\mathcal{X} \times \mathcal{Y})^n \times \mathcal{X}$ is universally measurable for every $n \in \mathbb{N} \cup \{0\}$. In words, a PAC learning algorithm is a function that maps a finite sequence of examples, and an instance, to a label.

Given a PAC learning algorithm $\mathbf{A}$, for every dataset $\mathcal{S} \in (\mathcal{X} \times \mathcal{Y})^{n}$, where $n \in \mathbb{N} \cup \{0\}$, define $\mathbf{A}_{\mathcal{S}} : \mathcal{X} \to \mathcal{Y}$ by
\begin{equation*}
    \mathbf{A}_{\mathcal{S}}(x) := \mathbf{A} \big( (\mathcal{S}, x) \big) \qquad \text{for all} \; x \in \mathcal{X}.
\end{equation*}

We note that for simplicity, the present work restricts attention to deterministic PAC learning algorithms. Nevertheless, our lower bounds continue to hold, up to constant factors, even when randomized PAC learning algorithms are allowed.

Afterward, we introduce three definitions: the first concerns the true error 
% \sbdcomment{What we are defining here is a \emph{loss} or, if you prefer \emph{true error} of a concept. There is no \emph{rate} involved since there is no process going on}
of a concept, the second concerns the empirical error of a concept, and the last concerns the error of a PAC learning algorithm.

We define the \emph{true error} of a concept $h$ with respect to a data distribution $\mathcal{D}$, denoted by $\mathcal{L}_{\mathcal{D}}(h)$, as follows:
\begin{equation*} \label{def: error-concept}
    \mathcal{L}_{\mathcal{D}}(h) := \mathbb{P}_{(x, y) \sim \mathcal{D}} \, \big[ h(x) \neq y \big].
\end{equation*}

Also, given $\mathcal{S} := \inbrace{ (x_1, y_1), (x_2, y_2) \ldots, (x_n, y_n)} \in (\mathcal{X} \times \mathcal{Y})^n$ for some $n \in \mathbb{N} \cup \{0\}$, we define \emph{empirical error} of a concept $h$ with respect to $S$, denote by $\mathcal{L}_{S}(h)$, as follows:
\begin{equation*} \label{def: emp-error}
    \mathcal{L}_{\cS}(h) := \frac{1}{n} \, \sum_{(x, y) \in S} \big[ h(x) \neq y \big].
\end{equation*}

Let $\mathbf{A}$ be a PAC learning algorithm. Given $\mathcal{S}$ as above, we define the \emph{error} of $\mathbf{A}_{\mathcal{S}}$ with respect to a data distribution $\mathcal{D}$, denoted by $\mathcal{L}_{\mathcal{D}}(\mathbf{A}_{\mathcal{S}})$, as follows: 
\begin{equation*} \label{def: error-algorithm}
    \mathcal{L}_{\mathcal{D}}(\mathbf{A}_{\mathcal{S}}) := \mathbb{P}_{(x, y) \sim \mathcal{D}} \, \big[ \mathbf{A}_{\mathcal{S}}(x) \neq y \big].
\end{equation*}

In this work, we restrict attention to deterministically labeled distributions. Formally, we have the following definition.

\begin{definition}[Deterministically labeled distribution] \label{def: dld}
A data distribution $\mathcal{D}$ is said to be \emph{deterministically labeled} if there exists a concept $f^{\star} : \mathcal{X} \to \mathcal{Y}$ such that
\begin{equation*}
    \mathcal{L}_{\mathcal{D}}(f^{\star}) = 0.
\end{equation*}
\end{definition}

Building on the previous definitions, we now define agnostic PAC learnability with deterministic labels. This definition is based on the same principles as those introduced by \cite{valiant1984theory, vapnik1974theory}.

\begin{definition}[Agnostic PAC learnability with deterministic labels] \label{def: pac-learnability}
We say that a concept class $\mathcal{C}$ is agnostic PAC learnable with deterministic labels, if there exist a PAC learning algorithm $\mathbf{A}$ and a sample complexity function $m^{\mathbf{A}}_{\mathcal{C}}: (0, 1) \times (0, 1) \to \mathbb{N}$ such that, for every deterministically labeled distribution $\mathcal{D}$, every $\varepsilon \in (0, 1)$, and every $\delta \in (0, 1)$, for $n \geq m^{\mathbf{A}}_{\mathcal{C}}(\varepsilon, \delta)$, we have:
\begin{equation*}
    \mathbb{P}_{\mathcal{S} \sim \mathcal{D}^n} \, \big[ \, \mathcal{L}_{\mathcal{D}}(\mathbf{A}_{\mathcal{S}}) \leq \, \inf_{h \in \mathcal{C}} \, \mathcal{L}_{\mathcal{D}}(h) +  \varepsilon \, \big] \geq 1 - \delta.
\end{equation*}

Now, let $\mathbf{A}$ be a PAC learning algorithm. We define the sample complexity of $\mathbf{A}$ as the function $\mathbf{m}_{\mathcal{C}}^{\mathbf{A}} : (0, 1) \times (0, 1) \to \mathbb{N}$ such that, for every $\varepsilon \in (0, 1)$ and $\delta \in (0, 1)$, the value $\mathbf{m}_{\mathcal{C}}^{\mathbf{A}}(\varepsilon, \delta)$ is the pointwise infimum over all functions satisfying the above guarantee for $\mathbf{A}$.

Finally, the optimal sample complexity of agnostic PAC learning $\mathcal{C}$ with deterministic labels, denoted by $\mathbf{m}_{\mathcal{C}}: (0, 1) \times (0, 1) \to \mathbb{N}$, is then defined as follows:
\begin{equation*}
    \mathbf{m}_{\mathcal{C}}(\varepsilon, \delta) := \inf_{\mathbf{A}} \, \mathbf{m}_{\mathcal{C}}^{\mathbf{A}}(\varepsilon, \delta) \qquad \text{for all} \; \varepsilon, \delta \in (0, 1).
\end{equation*}
that is, given $\varepsilon, \delta \in (0, 1)$, the minimal number of samples sufficient for the existence of some PAC learning algorithm that satisfies the above guarantee.
\end{definition}

\subsection{Combinatorial Complexity Parameters} \label{sec: pac-parameters}

In this subsection, we introduce the combinatorial complexity parameters appearing in our PAC learning results. We begin with the standard notions of shattering, VC dimension, and diameter. We then define two additional parameters, $\SETVC$ and $\PERTVC$, which are used in our analysis of the infinite-diameter regime.

\begin{definition}[Shattered set] \label{def: shattered-set}
Let $\mathcal{C} \subseteq \mathcal{Y}^\mathcal{X}$ be a concept class, and let $S \subseteq \mathcal{X}$ be a finite set of instances. We say that $S$ is \emph{shattered} by $\mathcal{C}$ if for every $f : S \to \{0, 1\}$, there exists a concept $c \in \mathcal{C}$ such that $c(x) = f(x)$ for all $x \in S$.
\end{definition}

\begin{definition}[$\operatorname{VC}$ dimension] \label{def: vc-dim}
Let $\mathcal{C} \subseteq \mathcal{Y}^\mathcal{X}$ be a concept class. The \emph{$\operatorname{VC}$ dimension} of $\mathcal{C}$, denoted by $\operatorname{VC}(\mathcal{C}) \in \mathbb{N} \cup \{0, +\infty\}$, is defined by the $\sup_{d \in \mathbb{N} \cup \{0\}}$ such that, there exists a finite set of instances $S \subseteq \mathcal{X}$ of size $d$ that is shattered by $\mathcal{C}$. Also, if $\mathcal{C} = \emptyset$, we have: $\operatorname{VC}(\mathcal{C}) = 0$.
\end{definition}

\begin{definition}[Diameter, following~\cite{ben2014sample}] \label{def: diameter}
Let $\mathcal{C} \subseteq \mathcal{Y}^{\mathcal{X}}$ be a concept class. The \emph{diameter} of $\mathcal{C}$, denoted by $\operatorname{D}(\mathcal{C}) \in \mathbb{N} \cup \{0, +\infty\}$, is defined by
\begin{equation*}
    \operatorname{D}(\mathcal{C}) := \sup_{h, h^{\prime} \in \mathcal{C}} \, | h \Delta h' |, \qquad h \Delta h' := \{ \, x \in \mathcal{X} \mid h(x) \neq h'(x) \, \}
\end{equation*}
Also, if $|\mathcal{C}| \leq 1$, we have: $\operatorname{D}(\mathcal{C}) = 0$.
\end{definition}

% \sbdcomment{I think that the parameters SET-VC and PERT-VC should also be defined here. It is odd to state results in terms of notions that are not defined in the main body of the pape}

The combinatorial parameter $\SETVC$ defined in \autoref{def:setVC} is adapted from the
set-shattering notion of \citet{ben2014sample}.
This notion was used in \citet{ben2014sample} to obtain lower bounds for agnostic PAC learning
with deterministic labels in the infinite-diameter regime.

\begin{definition}[\cite{ben2014sample}]
    Let $\cC$ be a concept class over a domain $\cX$, and let
    $A_1,\ldots,A_d \subseteq \cX$. We say that $\cC$ set-shatters
    $A_1,\ldots,A_d$ if, for every binary vector
    $\sigma=\inparen{\sigma_1,\ldots,\sigma_d} \in \{0,1\}^d$, there exists
    some $h_\sigma \in \cC$ such that, for every $i \leq d$ and every
    $x \in A_i$,
    \(
        h_\sigma(x)=\sigma_i.
    \)    
\end{definition}

\begin{definition} \label{def:setVC}
    For any concept class $\cC$ over domain $\cX$, define \emph{set VC dimension of $\cC$} denoted by $\SETVC(\cC)$ as the largest $d \in \Nat \cup \{0\}$ such that,
    for every $k \in \Nat$, there exist subsets
    $A_1,\ldots,A_d \subseteq \cX$, each of size at least $k$, that are
    set-shattered by $\cC$. If no finite largest such $d$ exists, we define
    $\SETVC(\cC)=\infty$.
\end{definition}

The combinatorial parameter $\PERTVC$ defined in \autoref{def:pertVC} measures the smallest VC
dimension that can be achieved after modifying each concept on at most a bounded number of points.
Intuitively, $\PERTVC(\cC,k)$ captures how simple the class can become after allowing $k$
perturbations per concept, while $\PERTVC(\cC)$ captures the limiting value of this quantity as the
perturbation budget grows.
In \autoref{sec: infinite-diameter}, we use this parameter to derive upper bounds for agnostic PAC
learning with deterministic labels in the infinite-diameter regime.

\begin{definition} \label{def:pertVC}
   Let $\cC$ be a concept class and let $k \in \Nat \cup \{0\}$. For every concept class $\cC'$, define the pseudo-distance from $\cC$ to $\cC'$ by
    \[
        d(\cC;\cC') :=  \inf \left\{ k : \forall h \in \cC \ \exists h' \in \cC' \text{ such that } |h \Delta h'| \leq k \right\}.
    \]
    Define \emph{$k-$perturbation VC dimension} of $\cC$ denoted by $\PERTVC(\cC, k)$ and \emph{perturbation VC dimension} of $\cC$ denoted by $\PERTVC(\cC)$ as
    \[
    \begin{aligned}
        \PERTVC(\cC,k) &:= \min\inbrace{ \VCD(\cC') \mid \cC' \text{ is a concept class such that } d(\cC;\cC') \leq k},
        \\
        \PERTVC(\cC) &:= \min\inbrace{ \VCD(\cC') \mid \cC' \text{ is a concept class such that } d(\cC;\cC') < \infty}.
    \end{aligned}
    \]
    % Denote $\cF_k$ be set of functions $f: \cC \rightarrow \{0, 1\}^{\cX}$ such that for every $h \in \cC$, $|f(h) \Delta h| \leq k$. Define 
    % \emph{$k-$perturbation VC dimension of $\cC$}, denote by 
    % $\PERTVC(\cC, k)$ as
    % \[
    % \PERTVC(\cC, k) := \inf \inbrace{\VCD(\cC_f) \mid f \in \cF_k }, \quad \text{ where } \cC_{f} := \{f(c) \mid c \in \mathcal{C}\}
    % \]

    % and define \emph{Perturbed VC dimension of $\cC$} denoted by $\PERTVC(\cC)$ as
    % \[
    % \PERTVC(\cC) := \lim_{k \rightarrow \infty} \PERTVC(\cC, k)
    % \]
\end{definition}

\section{Analysis of Two Concepts -- Finite Diameter} \label{sec: two-concepts}

In this section, we state our upper bounds for
optimal sample complexity of agnostic PAC learning for concept classes with two concepts. 

% \fmedit{This case serves as a toy example for the analysis developed in the remainder of the paper.}

Our analysis employs empirical risk minimization with memorization. For a concept class $\cC$ and a sample $\cS \in (\cX \times \cY)^*$, a predictor $h: \cX \rightarrow \cY$ is called an empirical risk minimizer (ERM) with respect to $\cC$ over $\cS$ if
\[
h \in \argmin_{h' \in \cC} \cL_\cS(h').
\]
Furthermore, for any concept $h$, define
\begin{equation} \label{eq:MERM}
    g_{h,\cS}(x) :=
\begin{cases}
y & \text{if } (x,y) \in \cS,\\
h(x) & \text{o.w.}
\end{cases}
\end{equation}
A predictor $f$ is called a memorizing empirical risk minimizer (MERM) with respect to $\cC$ over $\cS$ if there exists an empirical risk minimizer $h$ such that $f =g_{h,\cS}$. Furthermore we say that a PAC learner $\bA$ is ERM / MERM with respect to $\cC$, if for every $\cS \in  (\cX \times \cY)^*$, $\bA_\cS$ is ERM / MERM with respect to $\cC$ over $\cS$.

\begin{theorem} \label{thm:2-concept-upper}
    Let $K \in \mathbb N$. Any concept class with two concepts $\cC = \{h_0, h_1\}$ such that $D(\cC) \leq K$, is agnostic PAC learnable with deterministic labels with sample complexity upper bounded by 
    \[
    \mathbf m_\cC(\varepsilon, \delta) \leq \frac{8 \inparen{ \ln (4 / \delta)^{2 / 3} K^{1 / 3} +  2 \ln(4 / \delta)}}{\varepsilon} 
    \]
\end{theorem}
\begin{proof}
If, for a fixed instance, we flip the label assigned to that instance by every concept, our analysis remains unchanged. Therefore, without loss of generality we can replace $h_0$ with all-zero function and $h_1$ with $h_0 \Delta h_1$. This indicates that $|h_1| \leq K$.

Fix $\varepsilon,\delta\in(0,1)$ and fix an arbitrary deterministically
labeled distribution $\cD$ over $\cX\times\cY$. Let $\delta' = \nicefrac{\delta}{4}$, and define 
% $\OPT_{\cC}(\cD)$ as
\[\OPT_{\cC}(\cD) := \inf_{h \in \cC} \, \cL_{\cD}(h).\]

For any $x \in \cX$, denote $p_x = \cD\inbrace{(x, 0) \cup (x, 1)}$. Furthermore, let $q := \sum_{x \in h_1} p_x$ be the weight of disagreement of $h_0$ and $h_1$, and $\mu := \frac{\sum_{x \in h_1: f^*(x) = 1} p_x}{q}$ be the ratio of positive labels in the region of disagreement. In addition, without loss of generality assume that $\mu \geq \frac{1}{2}$. This implies that
$\cL_{\cD}(h_1) = \OPT_{\cC}(\cD)$. 

For any $n \in \Nat$, let $\cS$ be an i.i.d drawn sample of size $n$ from $\cD$.
Denote 
$n^+_\cS := |(x, y) \in \cS \mid x \in h_1|$ as number of instances of $\cS$ in $h_1$, and
$p^+_\cS := \sum_{x \in h_1: (x, 1) \in \cS} p_x$
as weight of positive instances memorized, and set
\[
\gamma := \frac{n^+_\cS \mu^2}{2 (K + n^+_\cS \mu)} + \frac{\varepsilon}{2 q}. \quad 
\]

We show that any PAC learner $\bA$ that is MERM with respect to $\cC$ satisfies
\[m_{\cC}^{\bA}(\varepsilon, \delta) \leq \frac{8 \inparen{ \ln (4 / \delta)^{2 / 3} K^{1 / 3} +  2 \ln(4 / \delta)}}{\varepsilon}.\]
Let $\hat h$ be a concept satisfying
$$\hat h \in \argmin_{h \in \cC} \cL_{\cS} (h), \quad \bA_\cS = g_{\hat h, \cS}.$$ 
Before proceeding to the proof, we state the following claims. The proof of Claim 2 is stated here and rest of them are deferred to Appendix~\ref{sec: upper-2-concepts}

\textbf{Claim 1:} Given a fixed $n^+_\cS$ we have  \[\prob{\frac{p^+_\cS}{q} < \frac{n^+_\cS \mu^2 }{K + n^+_\cS \mu} - \gamma} < \exp \inparen{-  n^+_\cS \gamma^2 }.\]

\textbf{Claim 2:} Given that $n^+_\cS \geq \frac{2 \ln(1 / \delta')}{\gamma^2}$ with probability at least $1 - 3 \delta'$ we have
$\cL_{\cD}(\bA_\cS) \leq \varepsilon.$

\textbf{Claim 3:} For any $K, n \in \mathbb N$ and $\varepsilon, q \in (0, 1]$ we have 
\[
F(q) := \frac{ q }{\inparen{\frac{nq^2 }{8 K + 2 nq} + \varepsilon }^2 } \leq \max \inbrace{ \frac{9}{4\sqrt{3}} \sqrt{\frac{K}{n}} \cdot \frac{1}{\varepsilon^{3/2}}, \frac{1}{\varepsilon}}
\]
\medskip

We show that if the claims are held, the proof is complete. Note that if $q < \varepsilon$, we get
\[
\cL_{\cD}(\bA_\cS) \leq \cL_{\cD}(\hat h) \leq \cL_{\cD}(h_0) < \cL_{\cD}(h_1) + \varepsilon, 
\]
which completes the proof. Therefore, we can assume that $q \geq \varepsilon$. First, observe that for $n \geq \frac{8 \ln(1 / \delta')}{\varepsilon}$ Multiplicative Chernoff Bound (\autoref{lem:mult-chern-bound}) implies
\begin{equation} \label{eq:2-conpcepts-upper-n-and-nS}
    \prob{n^+_\cS \leq \frac{nq}{2}} \leq \exp \inparen{-\frac{nq}{8}} \leq \delta'
\end{equation}

Then observe that $\gamma$ is an increasing function of $\mu$. Since $\mu \ge \frac{1}{2}$, this implies that 
\begin{equation} \label{eq:2-conpcepts-upper-gamma}
    \gamma \ge \gamma' := \frac{n^+_\cS}{8 K + 4 n^+_\cS} + \frac{\varepsilon}{2q}.
\end{equation}
Combining with Claim 2, as long as $n^+_\cS \geq  \frac{2 \ln(1 / \delta')}{\gamma'^2},$
with probability at least $1 - 3\delta'$ we have $\cL_{\cD}(\bA_\cS) \leq \varepsilon$. Then observe that $n^+_\cS - \frac{2 \ln(1 / \delta')}{\gamma'^2}$ is an increasing function w.r.t. $n^+_\cS$. Combining this fact with \autoref{eq:2-conpcepts-upper-gamma} and \autoref{eq:2-conpcepts-upper-n-and-nS} as long as
\begin{equation}\label{eq:2-conpcepts-upper-sup-q}
    \frac{nq}{2} \geq \max\inbrace{ \frac{4 q \ln(1 / \delta')}{\varepsilon}, \frac{2 \ln(1 / \delta')}{\inparen{\frac{nq }{16 K + 4 nq} + \frac{\varepsilon}{2q}}^2 }}
\end{equation}
with probability at least $1 - 4 \delta' = 1 - \delta$ we have $\cL_{\cD}(\bA_\cS) \leq \varepsilon$. Claim 3 implies that
\[
 \sup_{q \in (0, 1]}  \frac{1}{q} \frac{4 \ln(1 / \delta')}{\inparen{\frac{nq }{16 K + 4 nq} + \frac{\varepsilon}{2q}}^2 } = 16 \ln(1 / \delta') \sup_{q \in (0, 1]} \frac{ q }{\inparen{\frac{nq^2 }{8 K + 2 nq} + \varepsilon }^2 } \leq 4 \ln(1 / \delta') \max \inbrace{\frac{4}{\varepsilon},   \sqrt{\frac{27 K}{n \varepsilon^3 }} }.
\]
Combining this with \eqref{eq:2-conpcepts-upper-sup-q} completes the proof.

\textbf{Proof of Claim 2:}
Due to Claim 1 with probability at least $1 - \delta'$ it holds that
\begin{equation} \label{eq:2-conpcepts-upper-claim1}
    \frac{p^+_\cS}{q} \geq \frac{n^+_\cS \mu^2 }{K + n^+_\cS \mu} - \gamma.
\end{equation}
We consider two cases. \textbf{Case 1:} $2\mu - 1 > \gamma$. Since, $\bA_\cS = g_{\hat h, \cS}$, this implies
\begin{equation*}
\begin{aligned}
    \cL_{\cD}(\bA_\cS) & \leq  \cL_{\cD}(g_{h_0, \cS})\\
    & = \cL_{\cD}(h_0) - p^+_\cS \\
    & = \cL_{\cD}(h_1) + \inparen{\mu - (1- \mu)} q - p^+_\cS \\
    & = \OPT_{\cC}(\cD) + q \inparen{2\mu - 1 - \frac{p^{+}_\cS}{q}} \\
    & < \OPT_{\cC}(\cD) + q \inparen{2 \gamma  - \frac{n^+_\cS \mu^2 }{(K + n^+_\cS \mu)}} = \OPT_{\cC}(\cD) + \varepsilon.
\end{aligned}
\end{equation*}
This completes the proof of the Claim.

\textbf{Case 2:} $2\mu - 1 \geq \gamma$. Note that every instance of $\cS' = \inbrace{x \in h_1: \exists y \in \cY: (x, y) \in \cS}$ is an i.i.d. sample from $\cD( x \mid x \in h_1)$. Therefore,  for every $x \in \cS'$ the random variable $A_x := \mathbbm{1}\inbrace{f^*(x) = 1}$ is a Bernoulli random variable with parameter $\mu$.  Thus, using Hoeffding inequality (Theorem~\ref{thm:Hoeffding}) the probability of $\hat h = h_0$ is upper bounded by
\begin{equation*}
\begin{aligned}
\prob{\frac{|x \in \cS': f^*(x) = 1 |}{|\cS'|}  < \frac{1}{2} \bigm \vert n^+_\cS }  & \leq 2 \exp \inparen{- 2 n^+_\cS \inparen{\mu - \frac{1}{2}}^2 } \\
&\leq 2 \exp\inparen{\frac{-n^+_\cS \gamma^2}{2}}  \leq 2 \delta'
\end{aligned}
\end{equation*}
By a union bound with \autoref{eq:2-conpcepts-upper-claim1} the proof will be completed.

\end{proof}

\section{Conclusion, Discussion, and Future Directions} \label{sec: conclusion}

In this manuscript, we study the fundamental problem of agnostic PAC learning under deterministic labels. We make significant progress on the optimal sample complexity of this problem, substantially improving both the upper and lower bounds of the previous work. Technically, our upper bound is obtained by augmenting an empirical risk minimizer with memorization, and the analysis of this hybrid procedure differs considerably from known analyses of PAC learning algorithms, which may be of independent interest. This use of memorization in our results also sheds light on its role in machine learning. We then extend our investigation to the problem of online binary classification, where memorization again plays an important role. Eventually,
% in addition to improving our upper bound in Theorem~\ref{theorem: introduction_1} by a factor of $d^{1/3}$,
we discuss the following future directions:

\begin{itemize}[label=*]
    \item It would be interesting to investigate multiclass agnostic PAC learning under deterministic labels and agnostic PAC learning of partial concept classes under deterministic labels.

    \item Another natural direction for future work is to study agnostic active learning under deterministic labels. It is straightforward to see that, in the special case of concept classes with two functions, the optimal query complexity of active learning under deterministic labels is upper bounded by $K$ instead of $K^{1/3}/\varepsilon$. As a result, active learning may offer a significant advantage over passive learning in this setting. This direction is especially interesting to explore in light of the recent breakthrough result of \cite{hanneke2026agnostic}, whose proof uses the diameter in an essential way.

    \item A further natural direction for future work is to resolve the open problem concerning the relationship between $\SETVC(\cC)$ and $\PERTVC(\cC)$. In particular, it remains open whether these two parameters always coincide, or whether there exists a concept class $\cC \subseteq \{0, 1\}^{\mathcal{X}}$ for which $\SETVC(\cC) \neq \PERTVC(\cC)$. Resolving this question would clarify whether our upper bound in the infinite-diameter regime matches the existing lower bound in the small-$\varepsilon$ regime. 
\end{itemize}

\acksection

Steve Hanneke acknowledges support by grant no.\ 2024243 from the United States - Israel Binational Science Foundation (BSF). 
Shai  Ben-David’s  research  is  supported  by  an  NSERC Discovery  grant  and  through  a  CIFAR  AI  Chair.
Farnam Mansouri is supported by a NSERC Canada Graduate Scholarship-Doctoral (CGS-D), David
Cheriton Scholarship and a Vector Institute Research Grant.

\clearpage

\bibliographystyle{plainnat}
\bibliography{References}
\clearpage

\appendix

\section{Related Work} \label{sec: related-work}

\paragraph{Realizable PAC learning.}
The classical sample-complexity theory of PAC learning was initiated by
\citet{vapnik1974theory} and \citet{valiant1984theory}, and was later characterized in terms of
the VC dimension by \citet{blumer1989learnability}.
In the realizable setting, \citet{blumer1989learnability} showed that every class of VC dimension $d$ is learnable using empirical risk minimization (ERM), with sample complexity at most
\[
    \mathcal{O}\!\inparen{\frac{d\log(1/\varepsilon)+\log(1/\delta)}{\varepsilon}},
\]
and that $\Omega((d+\log(1/\delta))/\varepsilon)$ samples are necessary in general.
This upper bound is also known to be optimal for proper learners, i.e., learners whose outputs
belong to the concept class, including ERM
\cite{haussler1994predicting,auer2007new,simon2015almost,bousquet2020proper}.
For improper learners, the remaining logarithmic gap in the dependence on $\varepsilon$ was
resolved by \citet{hanneke2016optimal}, building on ideas of \citet{simon2015almost}, who proved
the optimal realizable PAC bound
\[
    \Theta\!\left(\frac{d+\log(1/\delta)}{\varepsilon}\right)
\]
for arbitrary VC classes.
More recently, there have been several attempts to prove optimal guarantees for other, arguably
simpler, learning algorithms.
These include the practical heuristic of bagging
\cite{larsen2023bagging, breiman1996bagging}, a variant of the one-inclusion graph learning
algorithm \cite{adenali2023optimal, haussler1994predicting}, and a simple majority vote among
three ERM classifiers \cite{aden2024majority}.

% \paragraph{ERM and proper learners.}
% The optimal learner introduced by \citet{hanneke2016optimal} is improper: it is constructed by taking
% a majority vote over carefully chosen classifiers from the concept class. This contrasts with the
% classical analysis of \citet{blumer1989learnability}, which applies to ERM learners. In fact, the
% extra $\log(1/\varepsilon)$ factor in the ERM bound is unavoidable in general; see, for instance,
% \citet{auer2007new}. The question of optimal sample complexity for proper learners was studied by
% \citet{bousquet2020proper}. \fmcomment{bad flow in general}

\paragraph{Agnostic PAC learning.}
In the agnostic setting, the first general VC-type bounds were obtained by
\citet{vapnik1974theory} and developed in the distribution-free decision-theoretic framework by
\citet{haussler1992decision}, showing that ERM attains sample complexity
\[
    \mathcal{O}\!\inparen{\frac{d\log(1/\varepsilon)+\log(1/\delta)}{\varepsilon^2}},
\]
and that $\Omega((d+\log(1/\delta))/\varepsilon^2)$ samples are necessary in general.
The extra $\log(1/\varepsilon)$ factor in the VC term was later removed using sharper
empirical-process techniques, most notably Talagrand's chaining bounds
\citep{10.1214/aop/1176988847}, showing that, unlike in the realizable setting, ERM achieves the
optimal agnostic PAC sample complexity
\[
    \Theta\!\left(\frac{d+\log(1/\delta)}{\varepsilon^2}\right).
\]

\section{Upper Bounds for Finite Diameter} 
\label{sec: upper}
In this section, we prove our upper bounds for optimal sample complexity of agnostic PAC learning with deterministic labels for concept classes with finite diameter. We first prove the two remaining claims used in the proof of \autoref{thm:2-concept-upper} in \autoref{sec: upper-2-concepts}. Then, in \autoref{sec: upper-genral-concepts} using the results for concept classes with two concepts, we prove our upper bounds for general VC classes.

\subsection{Missing Proofs for Two Concepts} \label{sec: upper-2-concepts}
In this subsection, we prove the two remaining claims used in the proof of \autoref{thm:2-concept-upper}. Before proving these claims, we recall an existing missing-mass concentration result, which allows us to convert bounds on the expectation of $p^+_\cS$ into high-probability bounds.

\begin{theorem}[Theorem 1 of \cite{10.1214/ECP.v18-2359}] \label{thm:miss-mass}
    Let $\mathbf{p}=\left(p_1, p_2, \ldots\right)$ be a distribution over $\mathbb{N}$ and suppose that $X_1, X_2, \ldots, X_n$ are sampled i.i.d according to $\mathbf{p}$. Define the indicator variable $\xi_j$ to be 0 if $j$ occurs in the sample and 1 otherwise:
    \[
    \xi_j=\mathbbm{1}_{\left\{j \notin\left\{X_1, \ldots, X_n\right\}\right\}}, \quad j \in \mathbb{N}.
    \]
    Denote the missing mass as the random variable
    $U_n=\sum_{j \in \mathbb{N}} p_j \xi_j.$ Then
    \[
    \prob{U_n > \expect{U_n} + \varepsilon} < \exp \inparen{- n \varepsilon^2}
    \]
\end{theorem}

We then proceed to the proof of the claims.

\begin{proofof}{Claim 1 of \autoref{thm:2-concept-upper}} We first bound expectation of $p^+_\cS$ given $n^+_\cS$ by
\begin{equation*}
\begin{aligned}
    \expect{p^+_\cS \mid n^+_\cS } & =  \sum_{x \in h_1: f^*(x) = 1} p_x \prob{(x, 1) \in \cS \mid n^+_\cS} \\
    & =  \sum_{x \in h_1: f^*(x) = 1} p_x \inparen{1 - \inparen{1 - \frac{p_x}{q}}^{n^+_\cS}} \\
    & \geq \sum_{x \in h_1: f^*(x) = 1} p_x \inparen{1 - e^{-\frac{p_x n^+_\cS}{q}}} \\
    & \geq \sum_{x \in h_1: f^*(x) = 1} p_x \frac{p_x n^+_\cS}{ q + n^+_\cS p_x} \\
    & = \sum_{x \in h_1: f^*(x) = 1}  \frac{p_x^2 n^+_\cS}{ q + n^+_\cS p_x} 
\end{aligned}
\end{equation*}
By applying Sedrakyan inequality (Lemma~\ref{lem:mult-chern-bound}) we get 
\begin{equation} \label{eq:2-conpcepts-upper-pS-exp}
    \begin{aligned}
        \expect{p^+_\cS \mid n^+_\cS} & \geq n^+_\cS \frac{\inparen{\sum_{x \in h_1: f^*(x) = 1} p_x}^2}{\sum_{x \in h_1: f^*(x) = 1} \inparen{q + n^+_\cS p_x} } \\
    & \geq n^+_\cS \frac{(q \mu)^2 }{K q + n^+_\cS q \mu } \\
    & = \frac{n^+_\cS q \mu^2 }{K + n^+_\cS \mu} \\
    % & \geq \frac{n^+_\cS q}{ 4 K + 2 n^+_\cS }
    \end{aligned}
\end{equation}
% Where the last line is due the fact that $\frac{n^+_\cS q \mu^2 }{K + n^+_\cS \mu}$ is an increasing function in $(0, 1]$ and $\mu \geq \frac{1}{2}$.

Observe that size of $V = \inbrace{x \in h_1, f^*(x) = 1}$ is at most $K$ and thus is countable. Consider probability distribution $\mathbf p$ over $U$ such that $\mathbf p(x) =\frac{p_x}{\mu q}$. Then, $ \cS':= \inbrace{x \in h_1: \exists y \in \cY: (x, y) \in \cS}$ would be i.i.d sample of size $n^+_\cS$ from $\mathbf p$.

Therefore, we can use Theorem~\ref{thm:miss-mass} over the domain $V$ and distribution $\mathbf p$, and i.i.d. samples $\cS'$. Then the missing mass $U_{n^+_\cS}$ would be $1 - \frac{p^+_\cS}{\mu q}$. By using \autoref{eq:2-conpcepts-upper-pS-exp} we get
\begin{equation*}
     \prob{\frac{p^+_\cS}{\mu q} < \frac{n^+_\cS \mu }{K + n^+_\cS \mu } - \gamma \Bigm \vert n^+_\cS} \leq \prob{U_{n^+_\cS} \geq \expect{U_{n^+_\cS} \mid n^+_\cS} + \gamma \Bigm \vert n^+_\cS}
     < \exp \inparen{- n^+_\cS \gamma^2 } 
\end{equation*}
Since $\mu \leq 1$ this implies that
\begin{equation} \label{eq:2-conpcepts-upper-pS-high-prob}
    \prob{\frac{p^+_\cS}{q} < \frac{n^+_\cS \mu^2 }{K + n^+_\cS \mu} - \gamma \Bigm \vert n^+_\cS} < \exp \inparen{-  n^+_\cS \gamma^2 } 
\end{equation}
This completes the proof of Claim 1.
\end{proofof}

\begin{proofof}{Claim 3 of \autoref{thm:2-concept-upper}}
We bound $\sup_{q \in (0, 1]} F(q)$ by considering two cases.

\medskip
\noindent
\textbf{Case 1: $nq \le 4K$.}
In this regime,
\(
8K + 2nq \le 16K,
\)
hence
\[
\frac{nq^2}{8K + 2nq} \ge \frac{nq^2}{16K}.
\]
Therefore,
\[
F(q) \le \frac{q}{\left(\varepsilon + \frac{nq^2}{16K} \right)^2}.
\]

Consider the function
\[
g(q) := \frac{q}{(\varepsilon + a q^2)^2}, \quad a := \frac{n}{16K}.
\]
A direct calculation shows that $g$ is maximized at
\(
q^2 = \frac{\varepsilon}{3a},
\)
and
\[
\max_{q > 0} g(q)
= \frac{9}{16\sqrt{3}} \cdot \frac{1}{\sqrt{a}\,\varepsilon^{3/2}}.
\]
Substituting $a = \frac{n}{16K}$ gives
\[
\sup_{q \in (0, 1]} F(q) \le \frac{9}{4\sqrt{3}} \sqrt{\frac{K}{n}} \cdot \frac{1}{\varepsilon^{3/2}}.
\]

\medskip
\noindent
\textbf{Case 2: $nq \ge 4K$.}
In this regime,
\(
8K + 2nq \le 4nq,
\)
so
\[
\frac{nq^2}{8K + 2nq} \ge \frac{q}{4}.
\]
Hence
\[
F(q) \le \frac{q}{(\varepsilon + q/4)^2}.
\]
The right-hand side is maximized at $q = 4\varepsilon$, yielding
\(
F(q) \le \frac{1}{\varepsilon}.
\)

\medskip
\noindent
Combining both cases completes the proof of claim 3.
\end{proofof}

\subsection{Analysis of General Concept Class}
\label{sec: upper-genral-concepts}

In this subsection section, we prove our upper bounds for agnostic PAC learning with deterministic labels of VC classes with finite diameter.

We first prove the result for finite concept classes. The proof reduces the problem to the two-concept case by applying the preceding two-concept guarantee uniformly over all pairs of concepts in $\cC$. Once this uniform event holds, an MERM over the full class can be compared to each competing concept $h \in \cC$ by restricting attention to the two-concept class consisting of $h$ and the ERM underlying the MERM predictor. This pairwise reduction is similar in spirit to ideas used in density estimation \citep{devroye2001combinatorial}.

\begin{theorem} \label{thm:N-concept-upper}
    Let $K \in \mathbb N$. There exists a universal constant $a_1 > 0$ such that every finite sized concept class $\cC$ such that $D(\cC) \leq K$, is agnostic PAC learnable with deterministic labels with sample complexity upper bounded by 
    \[
    \mathbf m_\cC(\varepsilon, \delta) \leq \frac{a_1 \inparen{ \ln \inparen{|\cC| / \delta}^{2 / 3} K^{1 / 3} +   \ln \inparen{|\cC| / \delta}}}{\varepsilon}
    \]
\end{theorem}
\begin{proof}
    Denote $M := \left(\begin{array}{c}|\cC|\\ 2\end{array}\right)$, and let $n \in \mathbb N$. Fix $\varepsilon, \delta \in (0, 1)$ and any deterministically labeled distribution $\cD$, and let $\cS$ be an i.i.d. sample of size $n$ drawn from $\cD$. Furthermore, suppose PAC learner $\bA$ is MERM with respect to $\cC$.

    Fix any two distinct concepts $h_1, h_2 \in \cC$. By \autoref{thm:2-concept-upper}, as long as 
    \[
    \begin{aligned}
        n \geq \frac{8 \inparen{ \ln (4 M / \delta)^{2 / 3} K^{1 / 3} +  2 \ln(4 M / \delta)}}{\varepsilon},
    \end{aligned}
    \] 
    every MERM $\bA^{h_1, h_2}$ with respect to $\inbrace{h_1, h_2}$, with probability at least $1 - \nicefrac{\delta}{M}$, satisfies
    \begin{equation} \label{eq:N-concepts-upper-2-concept}
        \cL_{\cD}(\bA^{h_1, h_2}_{\cS})  \leq \min_{i \in \{1, 2\}} \cL_{\cD}(h_i) + \varepsilon.
    \end{equation}
    Since $M \leq |\cC|^2$ and $\delta \in (0,1)$, it suffices for $n$ to satisfy
    \[
    n \geq \frac{16 \inparen{ \ln (2 |\cC| / \delta)^{2 / 3} K^{1 / 3} +  2 \ln(2 |\cC| / \delta)}}{\varepsilon}.
    \]
    By taking a union bound over all pairs of concepts in $\cC$, with probability at least $1 - \delta$, \autoref{eq:N-concepts-upper-2-concept} holds simultaneously for every pair of distinct concepts.

     Let $\hat h$ be any concept satisfying
     \[
     \hat h = \argmin_{h \in \cC} \cL_{\cS}(h), \quad \bA_\cS = g_{\hat h, \cS}
     \]
     Notice that for every $h \neq \hat h$ in $\cC$, $\bA_\cS$ is also MERM with respect to $\{h, \hat h\}$ over $\cS$.  Therefore, by \autoref{eq:N-concepts-upper-2-concept},
    \[
    \cL_{\cD}(\bA_{\cS}) \leq \min\inparen{\cL_{\cD}(h), \cL_\cD(\hat h)} + \varepsilon 
    \]
    Which implies that
    \[
    \cL_{\cD}(\bA_{\cS})  \leq \min_{h \in \cC} \cL_{\cD}(h) + \varepsilon 
    \]
    This completes the proof.
    
\end{proof}

We next extend the finite-class guarantee to VC classes. The standard way to pass from finite to infinite classes is to discretise the class on a suitably chosen sample. In our setting, the relevant discretization is obtained through an $\varepsilon$-net argument: after drawing an auxiliary sample from the distributions, concepts which their true error differ significantly must disagree on at least one point of this sample. This allows us to replace the original infinite class by a finite projection while preserving the comparisons needed for the finite-class theorem. Our descretization technique is similar to used in \cite{hopkins2022realizable}.

\begin{definition} [$\varepsilon-$net]
    Let $\cX$ be some domain, $\cB \subseteq 2^{\cX}$ a collection of subsets of $\cX$ and $\cQ$ a distribution over $\cX$. An $\varepsilon$-net for $\cW$ with respect to $\cQ$ is a subset $\cS \subseteq \cX$ that intersects every member of $\cB$ that has $\cQ$-weight at least $\varepsilon$.
\end{definition} 

\begin{lemma}[\cite{10.1145/10515.10522}] \label{lem:e-net}
    Let $\cC$ be a concept class over $\cX$ with $\VCD(\cC) = d < \infty$ and $\varepsilon, \delta \in (0,1)$. There exists a universal constant $a_2 > 0$ such that for every distribution $\cQ$ over $\cX$, a sample $\cS$ of size 
    \[
    n \geq \frac{a_2 \inparen{d \ln(1 / \varepsilon) + \ln(1 / \delta)}}{\varepsilon}
    \]
    i.i.d. drawn from $\cQ$ is $\varepsilon-$net for $\cC$ with respect to $\cQ$ with probability at least $1 - \delta$.
\end{lemma}

\begin{theorem} \label{thm:VC-upper}
    Let $K, d \in \mathbb N$. There exists a universal constant $a_3 > 0$ such that every concept class such that $\VCD(\cC) \leq d$ where $D(\cC) \leq K$, is agnostic PAC learnable with deterministic labels with sample complexity upper bounded by 
    \[\mathbf m_\cC(\varepsilon, \delta) \leq  \frac{ a_3 \inparen{\big(d\ln(1 / \varepsilon)+\ln(1/\delta)\big)^{2 / 3} \; K^{1 / 3} + d \ln(1 / \varepsilon)+ \ln(1 / \delta)}
    }{\varepsilon}.
    \]
    
\end{theorem}
\begin{proof} Throughout this proof, for every $T \in (\cX \times \cY)^*$ we denote the multiset $X(T) : =\{x \mid (x, y) \in T\}$. Also, for a $N \subseteq \cX$ and $h \in \cY^{\cX}$ we define restriction of $h$ to $N$ denoted by $h_{\mid N}$ as $\{h(x) \mid x \in N\}$.

Fix $\varepsilon,\delta\in(0,1)$ and fix an arbitrary deterministically
labeled distribution $\cD$ over $\cX\times\cY$.
% We prove that any MERM
% learner $\bA$ with respect to $\cC$, using sample $\cS\sim\cD^n$, achieves excess loss at most $\varepsilon$ with probability
% at least $1-\delta$.
Let 
\[
\begin{aligned}
    n_1 & = \frac{2a_2 \inparen{(2d + 1)\ln(2 / \varepsilon) + \ln(2 / \delta) }}{\varepsilon}, \\
    n_2 & = \frac{ 2a_1 \inparen{ \inparen{ d \ln(e n_1 / d) + \ln \inparen{4 / \delta}}^{2 / 3} K^{1 / 3} + d \ln(e n_1 / d) +  \ln \inparen{4 / \delta}}}{\varepsilon},
\end{aligned}
\]
where $a_1$ and $a_2$ are the universal constant defined in \autoref{thm:N-concept-upper} and \autoref{lem:e-net}. Let $\cS$ be an i.i.d sample of size $n \geq n_1 + n_2$ drawn from $\cD$. Denote $\cS_1$ as first $n_1$ element of $\cS$, and $\cS_2 = \cS \setminus \cS_1$.

Denote $\cC_{X(\cS_1)} := \inbrace{h|_{X(\cS_1)} : h\in\cC}$.  For each $h \in \cC_{X(\cS_1)}$, choose one representative concept
$h' \in \cC$ such that $h'|_{X(\cS_1)} = h|_{X(\cS_1)}$. Let the resulting finite subclass be denoted by
$\cC_1$. 
Let $\bA_\cS$ be MERM with respect to $\cC_1$ over $\cS_2$. It suffices to prove that with probability at least $1 - \delta$
\[
\cL_{\cD}(\bA_\cS) \leq \inf_{h \in \cC} \cL_{\cD}(h) + \varepsilon.
\]

\paragraph{$\cS_1$ is $\nicefrac{\varepsilon}{2}-$net.} First note that $\VCD(\cC \Delta \cC) \leq 2d + 1$ (it can be shown similar to the manner claim 1 of \cite{ben1998combinatorial} was proved). Thus, from \autoref{lem:e-net}, we have that $\cS_1$ is a $\nicefrac{\varepsilon}{2}-$net for $\cC \Delta \cC$ with respect to $\cD_{\cX}$ with probability at least $1 - \nicefrac{\delta}{2}$, where $\cD_{\cX}$ is the marginal distribution of $\cD$ over $\cX$. This indicates that, for every $h \in \cC$ and $h' \in \cC_1$ such that $h_{\mid X(\cS_1)} = h'_{\mid X(\cS_1)}$, we have $h \Delta h' \cap \cS_1 = \varnothing$. Thus, $\cD_{\cX}(h \Delta h') < \varepsilon / 2$. Therefore,
\[
    \cL_\cD(h) \leq \cL_{\cD}(h')  + \varepsilon/ 2,
\]
which implies that
\begin{equation} \label{eq:VC-upper-cover}
\inf_{h \in \cC} \cL_\cD(h) \leq \inf_{h' \in \cC_1} \cL_{\cD}(h')  + \varepsilon/ 2,
\end{equation}

\paragraph{Applying results for finite classes.} 
Notice that  \(
    D(\cC')\le D(\cC)\le K.\) Furthermore, by Sauer's Lemma \citep{sauer1972density, shelah1972combinatorial}, since $\VCD(\cC)\le d$,
    \begin{equation*} 
        |\cC_1| = 
        |\cC_{X(\cS_1)}|
        \le
        % \sum_{i=0}^d {n_1\choose i}.
        \left(\frac{e n_1}{d}\right)^d.
    \end{equation*}
    Thus, 
    \[
    \ln |\cC_1| \leq d \ln \inparen{\frac{e n_1}{d}}.
    \]
    Therefore,
    \[
    |\cS_2| \geq n_2
    \ge
    \frac{ 2a_1 \inparen{ \ln \inparen{4|\cC_1| / \delta}^{2 / 3} K^{1 / 3} +   \ln \inparen{4|\cC_1| / \delta}}}{\varepsilon}.
\]
Using \autoref{thm:N-concept-upper} for finite class $\cC_1$ with accuracy
parameter $\varepsilon/2$ and confidence parameter $\delta/2$, with probability at least $1-\nicefrac{\delta}{2}$ we have 
\[
    \cL_\cD(\bA_\cS)
    \le
    \inf_{h' \in\cC_1}\cL_\cD(h')+\varepsilon/2.
\]
Combining with \autoref{eq:VC-upper-cover} completes the proof.

% \paragraph{Discretization.} We first prove, by the probabilistic method, that there exists a finite
% subclass $\cC'\subseteq\cC$ which approximates $\cC$ in true loss.

% By the standard VC uniform convergence theorem (e.g. see \cite{haussler1992decision}) there exists a universal constant $a' > 1$ such that for
% \[n_1= \frac{a' d\ln\frac{1}{\varepsilon}}{\varepsilon^2}\]
% with probability at least $1/2$ over an i.i.d. sample
% \[ T= \inbrace{(x_1,y_1),\dots,(x_{n_1},y_{n_1})}\sim \cD^{n_1}, \]

\end{proof}

Combining our results with the
\( \mathcal{O}\!\inparen{\frac{K \ln(1/\varepsilon) + \ln(1/\delta)}{\varepsilon}} \)
upper bound of \cite{ben2009agnostic} for agnostic PAC learning
with deterministic labels for concept classes of diameter $K$, we immediately obtain the following
corollary.

\begin{corollary} \label{cor:VC-upper}
    Let $K, d \in \mathbb N$. There exists a universal constant $a_3 > 0$ such that every concept class such that $\VCD(\cC) \leq d$ where $D(\cC) \leq K$, is agnostic PAC learnable with deterministic labels with sample complexity upper bounded by 
    \[\mathbf m_\cC(\varepsilon, \delta) \leq \mathcal{O}\inparen{ \frac{  \inparen{\inparen{\big(d\ln(1 / \varepsilon)+\ln(1/\delta)\big)^{2 / 3} \; K^{1 / 3} + d\ln(1 / \varepsilon)}  \wedge K \ln(1 / \varepsilon)} + \ln(1 / \delta)}{\varepsilon}}.
    \]
    
\end{corollary}

\section{Lower Bounds} \label{sec: pac-lower}

In this section, we first introduce the additional information-theoretic notation used in the lower-bound arguments in \autoref{sec: information-notation}. We then prove our lower bounds for agnostic PAC learning under deterministic labels in \autoref{sec: lower-proof}.

\subsection{Information-Theoretic Notation} \label{sec: information-notation}

In this subsection, we collect a few standard information-theoretic notions and definitions used in the lower-bound proofs. Let $P$ and $Q$ be two probability measures on a common measurable space. If $P$ is absolutely continuous with respect to $Q$, we define the Kullback--Leibler divergence from $P$ to $Q$ by
\begin{equation*}
    \KL(P \| Q) := \int \log \left( \frac{dP}{dQ} \right) \, dP.
\end{equation*}
If $P$ is not absolutely continuous with respect to $Q$, we set
\begin{equation*}
    \KL(P \| Q) := +\infty.
\end{equation*}

Next, let $X$ and $Y$ be random variables defined on a common probability space. We define the mutual information between $X$ and $Y$ by
\begin{equation*}
    I(X;Y) := \KL(P_{X,Y} \| P_X \otimes P_Y),
\end{equation*}
where $P_{X,Y}$ denotes the joint distribution of $(X,Y)$, and $P_X$ and $P_Y$ denote the marginal distributions of $X$ and $Y$, respectively.

Finally, for $p \in [0, 1]$, we denote by $\operatorname{B}(p)$ the Bernoulli distribution with parameter $p$. That is, if $Z \sim \operatorname{B}(p)$, then
\begin{equation*}
    \mathbb{P}[Z = 1] = p \qquad \text{and} \qquad \mathbb{P}[Z = 0] = 1-p.
\end{equation*}

\subsection{Detailed Technical Overview of Lower Bounds} \label{sec:technical-overview-lower-bounds}

The principal difficulty is that, under deterministic labels, observing a point reveals its label. In particular, repeated occurrences of the same point provide no additional evidence about its label that can be exploited in a lower bound construction. Consequently, the hardness must be concentrated on a sufficiently large collection of points that remain unseen by the learner, who is unsure of their labels.

Both proofs place a small total probability mass $q$ uniformly over $k$ points and place the remaining mass on an anchor point at which the relevant concepts agree. The labels of the rare points are generated using slightly biased coins with a bias magnitude $\gamma$, but each label is sampled only once and is then fixed. Thus, every realization of the construction defines a genuinely deterministically labeled distribution. The randomization is used only as an auxiliary device in the lower bound argument, and a final application of the probabilistic method fixes one deterministic labeling on which the learner fails.

The parameters $q$ and $\gamma$ play complementary roles. The total mass of the rare region $q$ controls its importance, while $\gamma$ controls the advantage of the concept aligned with the hidden bias. We choose $q\gamma = C_2\epsilon$ for a sufficiently large universal constant $C_2$, so that predicting in the wrong biased direction on a constant fraction of the unseen rare points creates an excess error of order $q\gamma$ that is larger than $\epsilon$.

The proof must then balance two requirements. First, sufficiently many rare points must remain unseen. More specifically, let $M$ denote the number of distinct rare points observed in the sample. Since the learner may memorize the labels of these points, the excess error gap resulting from the relevant estimation error—an incorrect hidden sign in Lemma~\ref{sec: infinite-diameter} or errors on a constant fraction of the hidden coordinates in Lemma~\ref{lem:block-assouad-lower-bound}—has, schematically, the form
$$
\Omega(q\gamma) - O\left(\frac{qM}{k}\right).
$$
Therefore, to preserve a gap of order $q\gamma$, we require $M \in O(\gamma k)$. The quantity $M$ is at most the number of samples falling in the rare region, whose expectation is $nq$. Thus, imposing
$$
nq \lesssim \gamma k
$$
allows the corresponding concentration argument in each proof to ensure that $M \in O(\gamma k)$ holds with the required probability. Using $q = C_2\epsilon/\gamma$, this becomes
$$
\frac{n\epsilon}{\gamma} \lesssim \gamma k,
$$
or equivalently,
$$
n \lesssim \frac{k\gamma^2}{\epsilon}.
$$

Second, the sample must contain insufficient information to estimate the hidden parameter. Each distinct observed rare point contributes only $O(\gamma^2)$ to the relevant information quantity—KL divergence in Lemma~\ref{theorem: diameter-confidence-lower-bound} and mutual information in Lemma~\ref{lem:block-assouad-lower-bound}. Since the expected number of distinct observed rare points is at most $nq$, this information quantity is at most
$$
O(nq\gamma^2)
=
O(n\epsilon\gamma).
$$
To keep the corresponding estimation problem hard, we require
$$
n\epsilon\gamma \lesssim B,
$$
or equivalently,
$$
n \lesssim \frac{B}{\epsilon\gamma},
$$
where $B$ is the relevant information budget.

These two constraints can be summarized as
$$
n \lesssim \min \left \{ \frac{B}{\epsilon\gamma}, \frac{k\gamma^2}{\epsilon} \right \}.
$$
Balancing the two terms gives
$$
\gamma \asymp \left(\frac{B}{k}\right)^{1/3}, \qquad n \asymp \frac{k^{1/3}B^{2/3}}{\epsilon}.
$$

The two lower bounds instantiate this common mechanism with different hidden parameters and information budgets. In the principal regime $\log(1/\delta) \lesssim k$ of Lemma~\ref{theorem: diameter-confidence-lower-bound}, the hidden parameter is a single sign and the relevant information budget is $B = \log(1/\delta)$; a binary testing argument then yields the diameter-confidence term
$$
\frac{k^{1/3}\log^{2/3}(1/\delta)}{\epsilon}.
$$
The complementary regime and the additional confidence term are handled using the standard realizable PAC lower bound. In Lemma~\ref{lem:block-assouad-lower-bound}, the $k$ rare points are partitioned into $d$ blocks, the hidden parameter is a $d$-dimensional sign vector, and the relevant information budget is $B = d$; this yields the joint dependence
$$
\frac{k^{1/3}d^{2/3}}{\epsilon}.
$$

\subsection{Proof of the Lower Bounds} \label{sec: lower-proof}

\begin{theorem}[Lower bound based on diameter with confidence parameter] \label{theorem: diameter-confidence-lower-bound}
There exist universal constants $a, a_0, \delta_0 > 0$, and $k_0 \in \mathbb{N}$ such that the following holds. Let $k \geq k_0$, and let $\mathcal{C} \subseteq \mathcal{Y}^{\mathcal{X}}$ be a concept class. Assume that there exist two concepts $h^0, h^1 \in \mathcal{C}$ and distinct points
\begin{equation*}
    x_0, x_1, \ldots, x_k \in \mathcal{X}
\end{equation*}
such that $\{x_i\} \in \Sigma_{\mathcal{X}}$ for every $i \in \{0, 1, \ldots, k\}$, and
\begin{equation*}
    h^0(x_0) = h^1(x_0),
    \qquad
    h^0(x_i) \neq h^1(x_i) \;\; \text{for every} \;\; i \in [k].
\end{equation*}
Then, for every $\delta \in (0, \delta_0]$, writing
\begin{equation*}
    r_{\delta} := \log \left( \frac{1}{\delta} \right)
    \qquad \text{and} \qquad
    r_{\delta,k} := r_{\delta} \wedge k ,
\end{equation*}
and for every
\begin{equation*}
    \varepsilon \in \left( 0, \, a_0 \left( \frac{r_{\delta,k}}{k} \right)^{1/3} \right],
\end{equation*}
the optimal sample complexity of agnostic PAC learning $\mathcal{C}$ with deterministic labels satisfies
\begin{equation*}
    \mathbf{m}_{\mathcal{C}}(\varepsilon, \delta) \geq a \left( \dfrac{k^{1/3} \, r_{\delta,k}^{2/3}}{\varepsilon} + \frac{r_{\delta}}{\varepsilon} \right).
\end{equation*}
\end{theorem}

\begin{proof}
Throughout the proof, all constants are universal, and we decrease $a, a_0, \delta_0$ and increase $k_0$ whenever needed. As we defined in \autoref{sec: notation}, $\log(\cdot)$ is in base $2$, while some standard inequalities below are naturally written with base $\mathrm{e}$. Indeed, changing between the two bases only changes universal constants. In addition, for a data distribution $\mathcal{D}$, define $\operatorname{opt}_{\mathcal{C}}(\mathcal{D})$ as
\begin{equation*}
    \operatorname{opt}_{\mathcal{C}}(\mathcal{D}) := \inf_{h \in \mathcal{C}} \, \mathcal{L}_{\mathcal{D}}(h).\\
\end{equation*}

\circled{0} The main claim is that if $r_{\delta} \leq a_1 k$ for some constant $a_1 > 0$, then we have
\begin{equation*}
    \mathbf{m}_{\mathcal{C}}(\varepsilon,\delta) \geq a \frac{k^{1/3} \, r_{\delta}^{2/3}}{\varepsilon}.
\end{equation*}
Set
\begin{equation*}
    \gamma := C_1 \, \left( \frac{r_{\delta}}{k} \right)^{1/3},
\end{equation*}
where $C_1 > 0$ is a sufficiently large universal constant. By taking $a_1 > 0$ sufficiently small, we may assume $0 < \gamma \leq \frac{1}{8}$. Next, set
\begin{equation*}
        q := \frac{C_2 \, \varepsilon}{\gamma},
\end{equation*}
where $C_2 > 0$ is another sufficiently large universal constant. By taking $a_0 > 0 $ sufficiently small, the assumption $\varepsilon \leq a_0 ( r_{\delta}/{k} )^{1/3}$ implies that $0 < q \leq \frac{1}{2}$.\\

\circled{1} Define a marginal distribution $\mu$ over $\mathcal{X}$ by
\begin{equation*}
    \mu(x_0) := 1-q,
    \qquad \text{and} \qquad
    \mu(x_i) := \frac{q}{k} \;\; \text{for every} \;\; i \in [k].
\end{equation*}

Let the random variable $\sigma$ be uniformly distributed over $\{-1, +1\}$. Conditional on $\sigma$, let $Z_1, Z_2, \ldots, Z_k$ be independent Bernoulli random variables satisfying
\begin{equation*}
    \mathbb{P}[Z_i = 1 \mid \sigma = +1] = \frac{1}{2} + \gamma,
    \qquad \text{and} \qquad
    \mathbb{P}[Z_i = 1 \mid \sigma = -1] = \frac{1}{2} - \gamma.
\end{equation*}

For each realization of $(\sigma, Z_1, Z_2, \ldots, Z_k)$, define a deterministic labeling function $f_{\sigma,Z} : \mathcal{X} \to \mathcal{Y}$ by
\begin{equation*}
    f_{\sigma,Z}(x_0) := h^0(x_0) = h^1(x_0),
    \qquad \text{and} \qquad
    f_{\sigma,Z}(x_i) := h^{Z_i}(x_i) \;\; \text{for every} \;\; i \in [k], 
\end{equation*}
and define $f_{\sigma,Z}$ arbitrarily outside $\{x_0, x_1, \ldots, x_k\}$, for example by setting it equal to $h^0$ there. Since singletons $\{x_i\}$ are measurable, $f_{\sigma,Z}$ is measurable.

Now, for each realization of $(\sigma, Z_1, Z_2, \ldots, Z_k)$, let $\mathcal{D}_{\sigma,Z}$ be the distribution of
\begin{equation*}
    (x, f_{\sigma,Z}(x)),
\end{equation*}
where $x \sim \mu$. Indeed, $\mathcal{D}_{\sigma,Z}$ is deterministically labeled Definition~\ref{def: dld}.\\

\circled{2} Subsequently, fix a sample size $n$ that satisfies
\begin{equation*}
    n \leq a \frac{k^{1/3} \, r_{\delta}^{2/3}}{\varepsilon}.
\end{equation*}

Let $P_+$ and $P_-$ denote the laws of the sample $\mathcal{S} \sim \mathcal{D}_{\sigma,Z}^n$, averaged over $Z$, conditional on $\sigma = +1$ and $\sigma = -1$, respectively. Let $ I= (X_1, X_2, \ldots, X_n)$ denote the sequence of sampled instances and let $Y = (Y_1, Y_2, \ldots, Y_n)$ denote the corresponding sequence of labels. Now, by the chain rule for relative entropy, we have
\begin{equation*}
    \KL(P_+ \| P_-) = \KL(P_+^I \| P_-^I) + \mathbb{E}_{I \sim P_+^I} \, \left[ \KL \big( P_+^{Y \mid I} \, \big\| \, P_-^{Y\mid I} \big) \right],
\end{equation*}
where $P_\pm^I$ denote the marginals of $P_\pm$ on the instance sequence, and $P_\pm^{Y\mid I}$ denote the corresponding conditional laws of the label sequence given the instance sequence.
Since the marginal distribution of the sampled instances does not depend on
$Z$, we have $\KL(P_+^I \| P_-^I) = 0$. Therefore,
\begin{equation*}
    \KL(P_+ \| P_-) = \mathbb{E}_{I \sim P_+^I} \, \left[ \KL \big( P_+^{Y \mid I} \, \big\| \, P_-^{Y\mid I} \big) \right].
\end{equation*}

Now, fix a realization of $I$. Let $D(I)$ be the number of distinct points among $x_1, x_2, \ldots, x_k$ that appear in $I$. Conditional on $I$, only these $D(I)$ points can contribute to the relative entropy between the two conditional label distributions. Moreover, repeated appearances of the same point do not contribute additional independent information, because the label of each point is deterministic once $Z$ is fixed. Therefore, we have
\begin{equation*}
    \KL \big( P_+^{Y \mid I} \, \big\| \, P_-^{Y\mid I} \big)
    \leq
    D(I) \, \KL \bigg( \operatorname{B} \Big( \frac{1}{2} + \gamma \Big) \, \bigg| \, \operatorname{B} \Big( \frac{1}{2} - \gamma \Big) \bigg).
\end{equation*}
Note that, for $\gamma \leq 1/8$, we have
\begin{equation*}
    \KL \bigg( \operatorname{B} \Big( \frac{1}{2} + \gamma \Big) \, \bigg| \, \operatorname{B} \Big( \frac{1}{2} - \gamma \Big) \bigg) \leq 16 \, \gamma^2.
\end{equation*}
Thus, we have
\begin{equation*}
    \KL(P_+ \| P_-) \leq 16 \, \gamma^2 \, \mathbb{E}_{I \sim P_+^I} \, [D(I)].
\end{equation*}

Finally, $D(I) \leq N$, where $N$ is the total number of sampled points
falling in $\{x_1, x_2, \ldots, x_k\}$. Since $N \sim \operatorname{Bin}(n,q)$,
we have $\mathbb{E}[D(I)] \leq \mathbb{E}[N] = nq$. Therefore, we get
\begin{equation*}
    \KL(P_+ \| P_-) \leq 16 \, n \, q \, \gamma^2.
\end{equation*}

Now, using $q = C_2 \, \varepsilon/\gamma$ and the upper bound on $n$, we obtain
\begin{equation*}
    \KL(P_+ \| P_-) \leq 16 \, C_1 \, C_2 \, a \, r_{\delta}.
\end{equation*}
Taking $a > 0$ sufficiently small, the Bretagnolle--Huber inequality implies that every estimator $\widehat{\sigma} = \widehat{\sigma}(\mathcal{S})$ satisfies
\begin{equation*} \label{eq: testing-lower-bound}
    \mathbb{P}[\widehat{\sigma} \neq \sigma] \geq 8 \, \delta,
\end{equation*}
where the probability is over $\sigma$, $Z$, and the sample $\mathcal{S}$.\\

\circled{3} Next, fix a deterministic PAC learning algorithm $\mathbf{A}$. Now, we construct an estimator from the learner. For a sample $\mathcal{S}$, define
\begin{equation*}
    I(\mathcal{S}) := \{ i \in [k] \mid x_i \;\, \text{appears in} \;\, \mathcal{S}\},
    \qquad
    U(\mathcal{S}) := [k] \setminus I(\mathcal{S}),
\end{equation*}
and let $M(\mathcal{S}) := |I(\mathcal{S})|$. If
\begin{equation*}
    M(\mathcal{S}) > \frac{\gamma k}{16},
\end{equation*}
let $\widehat{\sigma}$ be arbitrary. Otherwise, define $\widehat{\sigma}$ as follows
\begin{equation*}
    \widehat{\sigma}(\mathcal{S})
    :=
    \begin{cases}
        +1,
        & \big| \{ i \in U(\mathcal{S}) \mid \mathbf{A}_{\mathcal{S}}(x_i) = h^1(x_i)\} \big| \geq |U(\mathcal{S})| / 2,\\
        -1,
        & \text{otherwise.}
    \end{cases}
\end{equation*}
Thus, in words, $\widehat{\sigma}$ is the majority vote of the learner's predictions on the unseen points.\\

\circled{4} We next show that if $\widehat{\sigma}$ is wrong, then the learner has excess error larger than $\varepsilon$, except on an event of probability at most $3\delta$.

\textbf{First}, let $N$ be the total number of sample points falling in $\{x_1, x_2, \ldots, x_k\}$. So, we have: $N \sim \operatorname{Bin}(n, q)$. Hence, we obtain
\begin{equation*}
    \mathbb{E}[N] = nq
    \leq
    a \cdot \frac{k^{1/3} \, r_{\delta}^{2/3}}{\varepsilon} \cdot \frac{C_2 \, \varepsilon}{\gamma}
    =
    \frac{a \, C_2}{C_1} \, k^{2/3} \, r_{\delta}^{1/3}. 
\end{equation*}
On the other hand, we have:
\begin{equation*}
    \frac{\gamma \, k}{16} = \frac{C_1}{16} \, k^{2/3} \, r_{\delta}^{1/3}. 
\end{equation*}
Taking $a > 0$ sufficiently small and noticing that $M(\mathcal{S}) \leq N$, a Chernoff bound gives
\begin{equation*}
    \mathbb{P} \left[ M(\mathcal{S}) > \frac{\gamma \, k}{16} \right] \leq \delta.\\
\end{equation*}

\textbf{Second}, define $h^{\sigma}$ as follows.
\begin{equation*}
    h^{\sigma}
    :=
    \begin{cases}
        h^1, & \sigma = +1,\\
        h^0, & \sigma = -1.
    \end{cases}
\end{equation*}
For every $i \in [k]$, the concept $h^{\sigma}$ disagrees with $f_{\sigma,Z}$ at $x_i$ with probability $1/2 - \gamma$. Therefore, we have:
\begin{equation*}
    \big| \{i \in [k] \mid h^{\sigma}(x_i) \neq f_{\sigma,Z}(x_i)\} \big| \sim \operatorname{Bin} \left( k, \frac{1}{2} - \gamma \right).
\end{equation*}
Since $h^{\sigma} \in \mathcal{C}$, Hoeffding's inequality gives
\begin{equation*}
    \mathbb{P}
    \left[ \, \operatorname{opt}_{\mathcal{C}}(\mathcal{D}_{\sigma,Z}) > q \left( \frac{1}{2} - \frac{3\gamma}{4} \right) \, \right]
    \leq
    \exp \left( - \frac{k\gamma^2}{8} \right)
    \leq
    \delta,
\end{equation*}
where the last inequality follows from the definition of $\gamma$ and from choosing $C_1$ large enough and $a_1$ small enough.\\

\textbf{Third}, condition on $\sigma$ and on a sample $\mathcal{S}$ satisfying
\begin{equation*}
    M(\mathcal{S}) \leq \frac{\gamma k}{16}
    \qquad \text{and} \qquad
    \widehat{\sigma}(\mathcal{S}) \neq \sigma.
\end{equation*}
Then, on at least half of the unseen indices $i \in U(\mathcal{S})$, the learner predicts the minority label relative to $\sigma$. Condition on $\sigma$ and $\mathcal{S}$, the labels on the unseen points remain independent. Hence, the expected loss contribution from unseen points is at least
\begin{equation*}
    q \left( 1 - \frac{\gamma}{16} \right) \, \frac{1}{2}.
\end{equation*}
As a result, by Hoeffding's inequality,
\begin{equation*}
    \mathbb{P}
    \left[
        \frac{q}{k}
        \sum_{i \in U(\mathcal{S})}
        \mathds{1}
        \left[
            \mathbf{A}_{\mathcal{S}}(x_i)
            \neq
            f_{\sigma,Z}(x_i)
        \right]
        <
        q \, \left( 1 - \frac{\gamma}{16} \right) \, \frac{1}{2}
        -
        \frac{q \, \gamma}{8}
        \;\middle|\;
        \sigma,\mathcal{S}
    \right]
    \leq
    \exp(-c \, k \, \gamma^2)
    \leq
    \delta,
\end{equation*}
for a universal constant $c > 0$, again by the choice of $C_1$ and $a_1$.\\

\textbf{Therefore}, except on an event of probability at most $3 \delta$, whenever $\widehat{\sigma}(\mathcal{S}) \neq \sigma$, we have
\begin{align*}
    \mathcal{L}_{\mathcal{D}_{\sigma,Z}}(\mathbf{A}_{\mathcal{S}}) - \operatorname{opt}_{\mathcal{C}}(\mathcal{D}_{\sigma,Z})
    & \geq
    q \, \left( 1 - \frac{\gamma}{16} \right) \, \frac{1}{2} - \frac{q \, \gamma}{8} - q \, \left( \frac{1}{2} - \frac{3\gamma}{4} \right)\\
    & =
    q \, \gamma \, \left( \frac{3}{4} - \frac{1}{32} - \frac{1}{8} \right)\\
    & =
    \frac{19}{32} \, q \, \gamma.
\end{align*}
Since $q \gamma = C_2 \, \varepsilon$ and $C_2$ is a sufficiently large universal constant, the last expression is larger than $\varepsilon$. Thus, we have:
\begin{equation*}
    \{ \widehat{\sigma}(\mathcal{S}) \neq \sigma\}
    \subseteq
    \left \{ \mathcal{L}_{\mathcal{D}_{\sigma,Z}}(\mathbf{A}_{\mathcal{S}}) > \operatorname{opt}_{\mathcal{C}}(\mathcal{D}_{\sigma,Z}) + \varepsilon \right\}
    \cup
    \mathcal{B},
\end{equation*}
where $\mathbb{P}[\mathcal{B}] \leq 3\delta$.\\

\circled{5} Afterward, combining the above result with the testing lower bound, namely \eqref{eq: testing-lower-bound}, gives
\begin{align*}
    \mathbb{P}_{\sigma,Z,\mathcal{S}}
    \left[
        \mathcal{L}_{\mathcal{D}_{\sigma,Z}}(\mathbf{A}_{\mathcal{S}})
        >
        \operatorname{opt}_{\mathcal{C}}(\mathcal{D}_{\sigma,Z})
        +
        \varepsilon
    \right]
    \geq
    \mathbb{P}[\widehat{\sigma} \neq \sigma] - 3 \delta
    \geq
    5 \delta.
\end{align*}
Now, we apply the probabilistic method. In particular, by averaging over $(\sigma,Z)$, there exists a fixed deterministic labeling $f_{\sigma,Z}$ such that
\begin{equation*}
    \mathbb{P}_{\mathcal{S}\sim\mathcal{D}_{\sigma,Z}^n}
    \left[
        \mathcal{L}_{\mathcal{D}_{\sigma,Z}}(\mathbf{A}_{\mathcal{S}})
        >
        \operatorname{opt}_{\mathcal{C}}(\mathcal{D}_{\sigma,Z})
        +
        \varepsilon
    \right]
    >
    \delta.
\end{equation*}
Since $\mathbf{A}$ was arbitrary, no algorithm using
\begin{equation*}
    n \leq a \frac{k^{1/3} \, r_{\delta}^{2/3}}{\varepsilon}
\end{equation*}
samples can satisfy the agnostic PAC learning with deterministic labels guarantee with confidence $1 - \delta$. Hence, whenever $r_{\delta} \leq a_1 k$, we have:
\begin{equation} \label{eq: lower_final}
    \mathbf{m}_{\mathcal{C}}(\varepsilon, \delta) \geq a \frac{k^{1/3} \, r_{\delta}^{2/3}}{\varepsilon}.\\
\end{equation}

\circled{6} Finally, if $r_{\delta} \leq a_1 k$, using \eqref{eq: lower_final} combined with the standard realizable PAC learning lower bound \cite{ehrenfeucht1989general}, we get
\begin{equation*}
    \mathbf{m}_{\mathcal{C}}(\varepsilon, \delta) \geq a \left( \frac{k^{1/3} \, r_{\delta}^{2/3}}{\varepsilon} + \frac{r_{\delta}}{\varepsilon} \right),
\end{equation*}
after decreasing $a$ by a universal factor.

Now, if $r_{\delta} > a_1 k$, again using the standard PAC learning lower bound \cite{ehrenfeucht1989general}, we get
\begin{equation*}
    \mathbf{m}_{\mathcal{C}}(\varepsilon, \delta) \geq a \frac{r_{\delta}}{\varepsilon}.
\end{equation*}

Since $r_{\delta} > a_1 k$, we have
\begin{equation*}
    k^{1/3} \, \big( r_{\delta} \wedge k \big)^{2/3} \leq a_1^{-1/3} r_{\delta}.
\end{equation*}
Thus, after decreasing $a$ once more, we have
\begin{equation*}
    \mathbf{m}_{\mathcal{C}}(\varepsilon, \delta) \geq a
    \left(
        \frac{k^{1/3} \, (r_{\delta} \wedge k)^{2/3}}{\varepsilon}
        +
        \frac{r_{\delta}}{\varepsilon}
    \right).
\end{equation*}
This proves the theorem.
\end{proof}

\begin{lemma}[Block version of the diameter lower bound] \label{lem:block-assouad-lower-bound}
There exist universal constants $a, a_0, \delta_0 > 0$ and $k_0 \in \mathbb{N}$ such
that the following holds. Let $d, k \in \mathbb{N}$ satisfy
\begin{equation*}
    1 \leq d \leq k,
    \qquad
    d \mid k,
    \qquad
    k \geq k_0.
\end{equation*}
Then, there exist a finite measurable space $(\mathcal{X}_{d,k}, 2^{\mathcal{X}_{d,k}})$ and a concept class $\mathcal{C}_{d,k} \subseteq \mathcal{Y}^{\mathcal{X}_{d,k}}$ such that
\begin{equation*}
    \operatorname{VC}(\mathcal{C}_{d,k})=d
    \qquad\text{and}\qquad
    \operatorname{D}(\mathcal{C}_{d,k})=k,
\end{equation*}
and, for every
\begin{equation*}
    \varepsilon \in \left( 0, \, a_0 \left( \frac{d}{k} \right)^{1/3} \right],
\end{equation*}
the optimal sample complexity of agnostic PAC learning with deterministic labels
satisfies
\begin{equation*}
    \mathbf{m}_{\mathcal{C}_{d,k}}(\varepsilon, \delta_0) \geq a \dfrac{k^{1/3} \, d^{2/3}}{\varepsilon}.
\end{equation*}
\end{lemma}

\begin{proof}
Throughout the proof, all constants are universal, and we decrease $a, a_0, \delta_0$ and increase $k_0$ whenever needed. As we defined in \autoref{sec: notation}, $\log(\cdot)$ is in base $2$, while some standard inequalities below are naturally written with base $\mathrm{e}$. Indeed, changing between the two bases only changes universal constants. In addition, for a data distribution $\mathcal{D}$, define $\operatorname{opt}_{\mathcal{C}}(\mathcal{D})$ as
\begin{equation*}
    \operatorname{opt}_{\mathcal{C}}(\mathcal{D}) := \inf_{h \in \mathcal{C}} \, \mathcal{L}_{\mathcal{D}}(h).\\
\end{equation*}

\circled{0} Set
\begin{equation*}
    \gamma := C_1 \, \left( \frac{d}{k} \right)^{1/3},
\end{equation*}
where $C_1 \in (0, 1/16)$ is a sufficiently small universal constant. Next, set
\begin{equation*}
    q := \frac{C_2 \, \varepsilon}{\gamma},
\end{equation*}
where $C_2 > 0$ is a sufficiently large universal constant. By taking $a_0 > 0$ sufficiently small, the assumption $\varepsilon \leq a_0 ( d/ k )^{1/3}$ implies that $ 0 < q \leq 1/2$.\\

\circled{1} Let $s = k / d$. Define $\mathcal{X}_{d,k}$ as 
\begin{equation*}
    \mathcal{X}_{d,k} := \{x_0\} \cup \{ x_{j,\ell} \mid j \in [d], \ell \in [s] \}
\end{equation*}
In addition, for every $v = (v_1, v_2, \ldots, v_d) \in \{-1, +1\}^d$, define $h^v:\mathcal{X}_{d,k} \to \{0,1\}$ by
\begin{equation*}
    h^v(x_0) := 0,
    \qquad \text{and} \qquad
    h^v(x_{j,\ell}) := \mathds{1}[v_j = +1] \;\; \text{for every} \;\; j \in [d], \ell \in [s].
\end{equation*}
Also, define $\mathcal{C}_{d,k}$ as
\begin{equation*}
    \mathcal{C}_{d,k} := \Big\{ h^v \mid v \in \{-1, +1\}^d \Big\}.\\
\end{equation*}

\circled{2} Next, we compute the combinatorial complexity parameters of $\mathcal{C}_{d,k}$. First, we analyze $\operatorname{VC}$ dimension. Observe that the set $\{x_{j,1} \mid j \in [d]\}$ is shattered by $\mathcal{C}_{d,k}$, so we have: $\operatorname{VC}(\mathcal{C}_{d,k}) \geq d$. Conversely, no set of size $d+1$ is shattered by $\mathcal{C}_{d,k}$. If it contains $x_0$, then the labeling assigns a value
$1$ to $x_0$ cannot be realized; otherwise, by the pigeonhole principle, it
contains two points from the same block $\{x_{j,\ell} \mid \ell \in [s]\}$, and every concept in $\mathcal{C}_{d,k}$ gives the same label to all points in this block. Thus, we have: $\operatorname{VC}(\mathcal{C}_{d,k}) < d+1$. Therefore, we have: $\operatorname{VC}(\mathcal{C}_{d,k}) = d$. Second, we analyze diameter. For $v, w \in \{-1, +1\}^d$, we have: $\big| \{ x \in \mathcal{X}_{d,k} \mid h^v(x) \neq h^w(x) \} \big| = s \big| \{ j \in [d] \mid v_j \neq w_j\} \big|$. Thus, taking $w = -v$ gives disagreement on all $k = ds$ non-anchor points. In addition, no pair can disagree on more points. Therefore, we have: $\operatorname{D}(\mathcal{C}_{d,k}) = k$.\\

\circled{3} Define a marginal distribution $\mu$ on $\mathcal{X}_{d,k}$ by
\begin{equation*}
    \mu(x_0) := 1-q,
    \qquad \text{and} \qquad
    \mu(x_{j,\ell}) := \frac{q}{k} \;\; \text{for every} \;\; j \in [d], \ell \in [s].
\end{equation*}

Let the random variable $V = (V_1, V_2, \ldots, V_d)$ be uniformly distributed over $\{-1, +1\}^d$. Conditional on $V$, let $Z_{j,\ell}$, $j \in [d]$, $\ell \in [s]$, be independent Bernoulli random variables satisfying
\begin{equation*}
    \mathbb{P}[Z_{j,\ell} = 1 \mid V] = \frac{1}{2} + \gamma \, V_j.
\end{equation*}

For each realization of $(V, Z)$, define a deterministic labeling function $f_{V,Z} : \mathcal{X}_{d,k} \to \{0, 1\}$ by
\begin{equation*}
    f_{V,Z}(x_0):=0, \qquad \text{and} \qquad f_{V,Z}(x_{j,\ell}) := Z_{j,\ell}.
\end{equation*}

Now, for each realization of $(V, Z)$, let $\mathcal{D}_{V,Z}$ be the distribution of
\begin{equation*}
    (x, f_{V,Z}(x)),
\end{equation*}
where $x \sim \mu$. Indeed, every $\mathcal{D}_{V,Z}$ is deterministically labeled Definition~\ref{def: dld}.\\

\circled{4} Subsequently, fix a sample size $n$ that satisfies
\begin{equation*}
    n \leq a \frac{k^{1/3} \, d^{2/3}}{\varepsilon},
\end{equation*}

Let $\mathcal{S} \sim \mathcal{D}_{V,Z}^n$. Let $I = (X_1, X_2, \ldots, X_n)$ denote the sequence of sampled instances, and let $Y = (Y_1, Y_2, \ldots, Y_n)$ denote the corresponding sequence of labels. By the chain rule for mutual information, we have
\begin{equation*}
    I(V;\mathcal{S}) = I(V;I,Y) = I(V;I) + \mathbb{E}_{I} \left[ I(V;Y \mid I) \right].
\end{equation*}

Since the marginal distribution of the sampled instances does not depend on
$V$, we have $I(V;I)=0$. Therefore,
\begin{equation*}
    I(V;\mathcal{S}) = \mathbb{E}_{I} \left[ I(V;Y \mid I) \right].
\end{equation*}

Now, fix a realization of $I$. Let $D(I)$ be the number of distinct points among $\{x_{j,\ell} \mid j\in[d],\, \ell\in[s]\}$ that appear in $I$. Conditional on $I$, only the labels of these $D(I)$ points can carry information about $V$. Moreover, repeated appearances of the same point do not contribute additional independent information, because the label of each point is deterministic once $V$ and $Z$ are fixed. Each such distinct observed point contributes at most $C \, \gamma^2$ information about the corresponding coordinate of $V$, for a universal constant $C>0$. Therefore, we have
\begin{equation*}
    I(V;Y\mid I) \leq C \, \gamma^2 \, D(I).
\end{equation*}
Thus, we have
\begin{equation*}
    I(V;\mathcal{S}) \leq C \, \gamma^2 \, \mathbb{E}[D(I)].
\end{equation*}

Finally, $D(I)\leq N$, where $N$ is the total number of sampled points falling in $\{x_{j,\ell} \mid j \in [d], \, \ell \in [s]\}$. Since $N \sim \operatorname{Bin}(n,q)$, we have $\mathbb{E}[D(I)] \leq \mathbb{E}[N] = nq$. Therefore, we get
\begin{equation*}
    I(V;\mathcal{S}) \leq C \, n \, q \, \gamma^2.
\end{equation*}

Now, using $q = C_2 \, \varepsilon/\gamma$ and the upper bound on $n$, we obtain
\begin{equation*}
    I(V;\mathcal{S}) \leq C \, C_2 \, a \, k^{1/3} \, d^{2/3} \, \gamma = C \, C_1 \, C_2 \, a \, d.
\end{equation*}
Taking $a > 0$ sufficiently small, we may assume that $I(V;\mathcal{S}) \leq c_0 d$, where $c_0 > 0$ is a sufficiently small universal constant.

By Assouad's lemma, every estimator $\widehat{V} = \widehat{V}(\mathcal{S})$ satisfies
\begin{equation*}
    \mathbb{E} \left[ \big| \{j\in[d] \mid \widehat{V}_j \neq V_j \} \big| \right] \geq \frac{d}{4}.
\end{equation*}
Consequently, we get
\begin{equation*}
    \mathbb{P}
    \left[
        \big| \{j\in [d] \mid \widehat{V}_j \neq V_j \} \big|
        \geq
        \frac{d}{8}
    \right]
    \geq
    \frac{1}{8}.
\end{equation*}

\circled{5} Next, fix a determinist PAC learning algorithm $\mathbf{A}$. Now, we construct an estimator from the learner. For every $j \in [d]$, define the unseen part of the $j$-th block by
\begin{equation*}
    U_j(\mathcal{S}) := \{ \ell \in [s] \mid x_{j,\ell} \;\, \text{does not appear in} \;\, \mathcal{S} \},
\end{equation*}
and let
\begin{equation*}
    M(\mathcal{S})
    :=
    \big|
        \{(j,\ell) \in [d] \times [s]
        \mid
        x_{j,\ell} \;\, \text{appears in} \;\, \mathcal{S}\}
    \big|
\end{equation*}
Define $\widehat{V}(\mathcal{S}) \in \{-1,+1\}^d$ by
\begin{equation*}
    \widehat{V}_j(\mathcal{S})
    :=
    \begin{cases}
        +1,
        &
        \big|
            \{\ell\in U_j(\mathcal{S})
            \mid
            \mathbf{A}_{\mathcal{S}}(x_{j,\ell})=1\}
        \big|
        \geq
        |U_j(\mathcal{S})|/2,
        \\
        -1,
        & \text{otherwise.}
    \end{cases}
\end{equation*}
Thus, if $\widehat{V}_j(\mathcal{S}) \neq V_j$, then the learner predicts against the biased direction on at least half of the unseen points in block $j$.

\circled{6} We now show that, on $\mathcal{E}$, the learner has an excess error larger than $\varepsilon$, except for an event with a small probability.

\textbf{First}, since $M(\mathcal{S})$ is at most the total number of samples falling outside $x_0$, we have
\begin{equation*}
    \mathbb{E}[M(\mathcal{S})] \leq nq.
\end{equation*}
Moreover, we have
\begin{equation*}
    nq \leq a \frac{k^{1/3} \, d^{2/3}}{\varepsilon} \cdot \frac{C_2 \, \varepsilon}{\gamma} = \frac{a \, C_2}{C_1} k^{2/3} \, d^{1/3},
\end{equation*}
while
\begin{equation*}
    \gamma \, k = C_1 \, k^{2/3} \, d^{1/3}.
\end{equation*}
Taking $a > 0$ sufficiently small, Markov's inequality gives
\begin{equation*}
    \mathbb{P}
    \left[
        M(\mathcal{S}) > \eta \, \gamma \, k
    \right]
    \leq
    \frac{1}{32},
\end{equation*}
where $\eta > 0$ is a sufficiently small universal constant. Therefore, with probability at least $1/8-1/32$, both
\begin{equation*}
    \big| \{j \in [d] \mid \widehat{V}_j(\mathcal{S})\neq V_j \} \big|
    \geq
    \frac{d}{8}
\end{equation*}
and
\begin{equation*}
    M(\mathcal{S})\leq \eta \, \gamma \, k
\end{equation*}
hold. Denote this event by $\mathcal{E}$.

\textbf{Second}, define $h^V\in\mathcal{C}_{d,k}$ to be the concept corresponding to the vector $V$. Since
\begin{equation*}
    h^V(x_{j,\ell})=\mathds{1}[V_j=+1],
\end{equation*}
we have:
\begin{equation*}
    \mathbb{E}
    \left[
        \mathcal{L}_{\mathcal{D}_{V,Z}}(h^V)
        \mid V
    \right]
    =
    q\left(\frac{1}{2} - \gamma\right).
\end{equation*}
Again, by Hoeffding's inequality, except with probability at most $\exp(-c' \, k \, \gamma^2)$, where $c'>0$ is a universal constant, we have:
\begin{equation*}
    \mathcal{L}_{\mathcal{D}_{V,Z}}(h^V)
    \leq
    q\left(\frac{1}{2} - \gamma\right)
    +
    \frac{q \, \gamma}{64}.
\end{equation*}
Since $h^V \in \mathcal{C}_{d,k}$, we get:
\begin{equation*}
    \operatorname{opt}_{\mathcal{C}_{d,k}}(\mathcal{D}_{V,Z})
    \leq
    \mathcal{L}_{\mathcal{D}_{V,Z}}(h^V).
\end{equation*}

\textbf{Third}, define
\begin{equation*}
    B(\mathcal{S},V)
    :=
    \big|
        \{(j,\ell)
        \mid
        \ell \in U_j(\mathcal{S}),
        \;
        \mathbf{A}_{\mathcal{S}}(x_{j,\ell})
        \neq
        \mathds{1}[V_j=+1]\}
    \big|.
\end{equation*}
If $\widehat{V}_j(\mathcal{S})\neq V_j$, then the learner predicts against the biased direction on at least half of the unseen points in block $j$. Hence, on $\mathcal{E}$, we have
\begin{equation*}
    B(\mathcal{S},V)
    \geq
    \frac{1}{2}
    \left(
        \frac{d}{8} \cdot s
        -
        M(\mathcal{S})
    \right)
    \geq
    \frac{k}{16}
    -
    \frac{\eta \, \gamma \, k}{2}.
\end{equation*}

Conditional on $V$ and $\mathcal{S}$, the labels on the unseen points remain independent. Let
\begin{equation*}
    L_{\operatorname{un}}
    :=
    \frac{q}{k}
    \sum_{j=1}^{d}
    \sum_{\ell \in U_j(\mathcal{S})}
    \mathds{1}
    \left[
        \mathbf{A}_{\mathcal{S}}(x_{j,\ell})
        \neq
        f_{V,Z}(x_{j,\ell})
    \right].
\end{equation*}
Then, we have:
\begin{equation*}
    \mathbb{E}
    \left[
        L_{\operatorname{un}}
        \mid V,\mathcal{S}
    \right]
    =
    \frac{q}{k}
    \left[
        (k-M(\mathcal{S}))
        \left( \frac{1}{2} - \gamma \right)
        +
        2 \, \gamma \, B(\mathcal{S},V)
    \right].
\end{equation*}
Thus, by Hoeffding's inequality, except with probability at most $\exp(-c \, k \, \gamma^2)$, where $c>0$ is a universal constant, we have
\begin{equation*}
    L_{\operatorname{un}}
    \geq
    \mathbb{E}
    \left[
        L_{\operatorname{un}}
        \mid V,\mathcal{S}
    \right]
    -
    \frac{q \, \gamma}{64}.
\end{equation*}

\textbf{Therefore}, except with probability at most $\exp(-c k \, \gamma^2) + \exp(-c'\, k \, \gamma^2)$, on the event $\mathcal{E}$ we have
\begin{align*}
    \mathcal{L}_{\mathcal{D}_{V,Z}}(\mathbf{A}_{\mathcal{S}}) - \operatorname{opt}_{\mathcal{C}_{d,k}}(\mathcal{D}_{V,Z})
    & \geq L_{\operatorname{un}} - \mathcal{L}_{\mathcal{D}_{V,Z}}(h^V)\\
    & \geq
    \frac{q}{k}
    \left[
        -M(\mathcal{S})
        \left( \frac{1}{2} - \gamma \right)
        +
        2 \, \gamma \, B(\mathcal{S},V)
    \right]
    -
    \frac{q \, \gamma}{32}
    \\
    & \geq
    q \, \gamma
    \left(
        \frac{3}{32}
        -
        \frac{\eta}{2}
    \right).
\end{align*}
Taking $\eta > 0$ sufficiently small, the last expression is at least $c'' \, q \, \gamma$ for a universal constant $c'' > 0$. Since $q \, \gamma = C_2 \, \varepsilon$, choosing $C_2 > 0$ sufficiently large gives
\begin{equation*}
    c'' \, q \, \gamma > \varepsilon.\\
\end{equation*}

\circled{7} Finally, observe that
\begin{equation*}
    k\gamma^2 = C_1^2 \, k^{1/3} \, d^{2/3}.
\end{equation*}
Since $d \geq 1$, taking $k_0$ sufficiently large makes
\begin{equation*}
    \exp(-c \, k \, \gamma^2) + \exp(-c^{\prime} \, k \, \gamma^2)
\end{equation*}
smaller than a sufficiently small universal constant. Combining this with
$\mathbb{P}[\mathcal{E}] \geq 1/8-1/32$, we get that, for some universal constant $\delta_0 > 0$,
\begin{equation*}
    \mathbb{P}_{V,Z,\mathcal{S}}
    \left[
        \mathcal{L}_{\mathcal{D}_{V,Z}}(\mathbf{A}_{\mathcal{S}})
        >
        \operatorname{opt}_{\mathcal{C}_{d,k}}(\mathcal{D}_{V,Z})
        +
        \varepsilon
    \right]
    >
    \delta_0.
\end{equation*}
Now, we apply the probabilistic method. In particular, by averaging over $(V,Z)$, there exists a fixed realization $(v,z)$ such that, for the corresponding deterministically labeled distribution $\mathcal{D}_{v,z}$,
\begin{equation*}
    \mathbb{P}_{\mathcal{S}\sim\mathcal{D}_{v,z}^n}
    \left[
        \mathcal{L}_{\mathcal{D}_{v,z}}(\mathbf{A}_{\mathcal{S}})
        >
        \operatorname{opt}_{\mathcal{C}_{d,k}}(\mathcal{D}_{v,z})
        +
        \varepsilon
    \right]
    >
    \delta_0.
\end{equation*}
Since $\mathbf{A}$ was arbitrary, no algorithm using
\begin{equation*}
    n \leq a \frac{k^{1/3} \, d^{2/3}}{\varepsilon}
\end{equation*}
samples satisfies the agnostic PAC learning with deterministic labels guarantee with confidence $1-\delta_0$. Hence, we have:
\begin{equation*}
    \mathbf{m}_{\mathcal{C}_{d,k}}(\varepsilon,\delta_0) \geq a \frac{k^{1/3} \, d^{2/3}}{\varepsilon}.
\end{equation*}
This proves the theorem.
\end{proof}

\section{Analysis of Infinite Diameter} \label{sec: infinite-diameter}

In this section, we prove an upper bound for concept classes with infinite diameter. Note that the ``Outer $\cC^*$ approximation'' portion of the proof is inspired by upper bound of \citet{ben2014sample} for agnostic PAC learning with deterministic labels of classes with finite diameter.

\begin{theorem} \label{thm:infinite-dim-upper}
    For every $k \in \Nat$ and concept class $\cC$ with $d := \VCD(\cC) < \infty$, $\cC$ is agnostic PAC learnable with sample complexity upper bounded by

    %\sbdedit{I think that we do not need any $k$ or "universal constant $a$" and the bound should be restated with just $\PERTVC(\cC)$ replacing $\PERTVC(\cC, k)$ }
    \begin{equation} \label{eq:infinite-dim}
        \begin{aligned}
        \mathbf{m}_{\cC}(\varepsilon, \delta) & = \mathcal{O} \inparen{ \frac{ \PERTVC(\cC, k) + \ln(1 / \delta)}{\varepsilon^2} + \frac{k \ln(1 / \varepsilon)}{\varepsilon}} 
    \end{aligned}
    \end{equation}
\end{theorem}
\begin{proof}
    Let 
    \[
     \cC^* = \argmin_{\cC'} \{\VCD(\cC'): d(\cC; \cC') \leq k\}
    \]
    Fix $\varepsilon,\delta\in(0,1)$ and fix an arbitrary deterministically labeled distribution $\cD$ over $\cX\times\cY$. Let $n \in \Nat$, and $\cS$ be an i.i.d. sample of size $n$ drawn from $\cD$. Let $\bA$ be an MERM with respect to $\cC^*$.

    \paragraph{Outer $\cC^*$ approximation.} 
    We call a point $x \in \cX$ heavy if its marginal weight is at least $\nicefrac{\varepsilon}{2k}$. Let $T$ be the set of heavy points. Since each point in $T$ has marginal weight at least $\nicefrac{\varepsilon}{2k}$, we have $|T| \leq \frac{2k}{\varepsilon}$. Therefore, as long as 
    \[
    n \geq \frac{2k \ln(1 / \varepsilon) + 2\ln(2 / \delta)}{\varepsilon},
    \]
    with probability at least $1 - \nicefrac{\delta}{2}$, the sample $\cS$ contains every point in $T$. On this event, for every $h \in \cC^*$ and every $h' \in \cC$ such that $|h \Delta h'| \leq k$, we have
    \begin{equation} \label{eq:infinie-dim-k-ham}
        \cL_{\cD}(g_{h, \cS}) - \cL_{\cD}(h')
        \leq
        \cD_{\cX} \big(h \Delta h' \setminus \cS\big)
        \leq
        k \cdot \frac{\varepsilon}{2k}
        =
        \frac{\varepsilon}{2},
    \end{equation}
    where $\cD_{\cX}$ is the marginal distribution of $\cD$.

    \paragraph{Inner $\cC^*$ approximation.} Let $\hat h$ be a concept satisfying
    \[ \hat h \in \argmin_{h \in \cC^*} \cL_{\cS}(h), \quad \bA_\cS = g_{\hat h, \cS}\]
    By standard PAC theory (e.g. \cite{haussler1992decision}), there exists a universal constant $a > 0$ such that for
    \[
    n \geq  \frac{ a\inparen{\PERTVC(\cC, k)  + \ln(1 / \delta)}}{\varepsilon^2}
    \]
    with probability at least $1 - \nicefrac{\delta}{2}$
    \begin{equation*} 
        \cL_{\cD}(\hat h) \leq \inf_{h \in \cC^*} \cL_{\cD}(h) + \varepsilon / 2.
    \end{equation*}
    Note that for every $h, h' \in \cC^*$
    \[
    \cL_{\cD}(g_{h, \cS}) - \cL_{\cD}(g_{h', \cS}) \leq \cL_{\cD}(h) - \cL_{\cD}(h'),
    \]
    and that $\cL_{\cD}(g_{h, \cS}) \leq \cL_{\cD}(h)$. Thus, with probability at least $1 - \nicefrac{\delta}{2}$
    \begin{equation*} 
        \cL_{\cD}(g_{\hat h, \cS}) \leq \inf_{h \in \cC^*} \cL_{\cD}(g_{\hat h, \cS}) + \varepsilon / 2.
    \end{equation*}
    Combining this with \autoref{eq:infinie-dim-k-ham} with probability at least $1 - \delta$ we get
       \[
       \cL_{\cD}(g_{\hat h, \cS}) \leq \inf_{h \in \cC^*} \inf_{h' \in \cC: |h' \Delta h| \leq k} \cL_{\cD}(h') + \varepsilon \leq \inf_{h' \in \cC} \cL_{\cD}(h') + \varepsilon
       \]
    This completes the proof.
    
\end{proof}

\paragraph{Discussion.} First, recall the lower bound of \citet{ben2014sample} for agnostic PAC learning with deterministic labels in the infinite-diameter regime.

\begin{theorem}[\cite{ben2014sample}, restated in terms of $\SETVC$] \label{thm:infinite-dim-lower}
    For every concept class $\cC$ such that $\SETVC(\cC) < \infty$, the optimal sample complexity of agnostic PAC learning $\cC$ with deterministic labels satisfies
    \[
        \mathbf{m}_{\cC}(\varepsilon, \delta) \geq \frac{\SETVC(\cC)}{\varepsilon^2}
    \]
\end{theorem}

 Then observe that  \(\PERTVC(\cC)=\lim_{k\to\infty}\PERTVC(\cC,k).\)
Since $\PERTVC(\cC,k)$ is integer-valued and non-increasing in $k$, there exists
$k^\star\in\Nat$ such that
\[
    \PERTVC(\cC,k^\star)=\PERTVC(\cC).
\]
Fix such a $k^\star$.
The $1/\varepsilon^2$ term in our upper bound in \autoref{eq:infinite-dim} is controlled by
$\PERTVC(\cC)$.
Therefore, as $\varepsilon\to0$, this term dominates the lower-order term.
In particular, if $\SETVC(\cC)=\PERTVC(\cC)$, then our upper bound is tight in the
small-$\varepsilon$ regime.
This leads to the following natural open question.

\SetPertb*

% answering the following question determines 

% It is natural to ask whether this upper bound is tight up to logarithmic factors in the small-$\varepsilon$ regime.
 
 % Therefore, $\PERTVC(\cC)$ provides an upper bound the quadratic dependence of sample complexity of agnostic PAC learning with deterministic labels in infinite diameter regime in term of $1 / \varepsilon$ up to a $\ln(1 / \varepsilon)$ factor. Thus, a natural question would be that how tight is $\PERTVC(\cC)$?

% \input{Appendix/proper}

\section{Deferred Definitions} \label{sec: deferred-definitions}

In this section, we define the online learning framework in \autoref{sec: online-framework}. Then in \autoref{sec: online-parameters}, we give the definition of the main combinatorial complexity parameter that appears in our online learning results in \autoref{sec: online-results}.

\subsection{Online Learning Framework} \label{sec: online-framework}

We begin by giving the problem setup. Fix a non-empty set $\mathcal{X}$, referred to as the instance space. In addition, fix $\mathcal{Y} = \{0, 1\}$ as the label space. A concept is a function from $\mathcal{X}$ to $\mathcal{Y}$. With this in mind, fix a non-empty collection of such functions $\mathcal{C}$ as the concept class. We specify instances of the online learning framework using the concept class $\mathcal{C}$.

Next, fix $\operatorname{T} \in \mathbb{N}$. The problem of online learning is formulated as a $\operatorname{T}$-rounded sequential game between a learner and an adversary. In particular, on each round $t \in [\operatorname{T}]$:

\begin{enumerate}
    \item The adversary chooses an instance $x_t \in \mathcal{X}$ and presents it to the learner.
    \item The adversary chooses a label $y_t \in \mathcal{Y}$.
    \item The learner predicts a label $\hat{y}_t \in \mathcal{Y}$.
    \item The adversary reveals the label $y_t$ to the learner.
\end{enumerate}

Subsequently, we present the definition of an online learning algorithm. Let $\mathcal{D} = \big\{ (\text{T}, \text{S}, \text{x}) \mid \text{T} \in \mathbb{N}, \, \text{S} \in \cup_{t = 0}^{\text{T} - 1} (\mathcal{X} \times \mathcal{Y})^{t}, \, \text{x} \in \mathcal{X} \big \}$. A randomized online learning rule is a mapping from $\mathcal{D}$ to $\Pi(\mathcal{Y})$, where $\Pi(\mathcal{Y})$ denotes the set of all probability distributions over $\mathcal{Y}$. In words, an online learning algorithm is a function that maps a time horizon, a finite sequence of instances of length at most the time horizon, and an instance, to a distribution over the labels.

Fix an online learning algorithm $\mathbf{A}$. Also, fix $\operatorname{T} \in \mathbb{N}$. Let $\mathcal{S}$ be a finite sequence of instance label pairs of size $\operatorname{T}$, that is, $\mathcal{S} \in (\mathcal{X} \times \mathcal{Y})^{\operatorname{T}}$. We denote by $\mathbf{A}(\mathcal{S}) \in \Pi(\mathcal{Y})^{\operatorname{T}}$ the random sequence of outputs of $\mathbf{A}$ when the adversary's choices are obtained according to $\mathcal{S}$.

In this work, we restrict attention to deterministically labeled sequences. Formally, we have the following definition.

\begin{definition}[Deterministically labeled sequence] \label{def: dls}
A sequence of instance-label pairs $\mathcal{S} = \big( (x_1,y_1), (x_2, y_2), \ldots, (x_{\operatorname{T}}, y_{\operatorname{T}}) \big) \in (\mathcal{X} \times \mathcal{Y})^{\operatorname{T}}$ for some $\operatorname{T} \in \mathbb{N}$ is said to be \emph{deterministically labeled} if there exists a concept $f^{\star} : \mathcal{X} \to \mathcal{Y}$ such that
\begin{equation*}
    f^{\star}(x_t) = y_t \qquad \text{for all} \;\; t \in [T].
\end{equation*}
\end{definition}

To assess the performance of online learning algorithms, we adapt the well-established game-theoretic notion of expected regret to our framework.

\begin{definition}[Expected Number of Mistakes] \label{def: expected-number-of-mistakes}
Let $\mathbf{A}$ be a randomized online learning rule. Fix $\operatorname{T} \in \mathbb{N}$. Let $\mathcal{S}$ be a finite sequence of instance-label pairs of size $\operatorname{T}$, namely $\mathcal{S} = \big( (x_1, y_1), (x_2, y_2), \dots, (x_{\operatorname{T}}, y_{\operatorname{T}}) \big) \in (\mathcal{X} \times \mathcal{Y})^{\operatorname{T}}$. We define the expected total number of mistakes made by $\mathbf{A}$ on the sequence $\mathcal{S}$, denoted by $\widebar{\mathbf{M}}(\mathbf{A}; \mathcal{S})$, as follows:
\begin{equation*}
    \widebar{\mathbf{M}}(\mathbf{A}; \mathcal{S}) := \mathbb{E} \, \bigg [ \sum_{t = 1}^{\operatorname{T}} \, \mathds{1} [ y_t \neq \hat{y}_t ] \bigg ],
\end{equation*}
where $(\hat{y}_1, \hat{y}_2, \dots, \hat{y}_{\operatorname{T}}) \sim \mathbf{A}(\mathcal{S})$.
\end{definition}

\begin{definition}[Expected Regret] \label{def: expected-regret}
Let $\mathbf{A}$ be a randomized online learning rule. Fix $\operatorname{T} \in \mathbb{N}$. Let $\mathcal{S}$ be a finite sequence of instance-label pairs of size $\operatorname{T}$, namely $\mathcal{S} = \big( (x_1, y_1), (x_2, y_2), \dots, (x_{\operatorname{T}}, y_{\operatorname{T}}) \big) \in (\mathcal{X} \times \mathcal{Y})^{\operatorname{T}}$. We define the expected regret of $\mathbf{A}$ on the sequence $\mathcal{S}$ against the performance of the concept class $\mathcal{C}$, denoted by $\mathbf{R}(\mathbf{A}; \mathcal{S}; \mathcal{C})$, as follows:
\begin{equation*}
    \mathbf{R}(\mathbf{A}; \mathcal{S}; \mathcal{C}) := \widebar{\mathbf{M}}(\mathbf{A}; \mathcal{S}) - \inf_{c \in \mathcal{C}} \, \sum_{t = 1}^{\operatorname{T}} \, \mathds{1} \big [ c(x_t) \neq y_t \big ].
\end{equation*}
\end{definition}

Finally, we define the best achievable worst-case expected regret in our framework.

\begin{definition}[Optimal Expected Regret with deterministic labels] \label{def: optimal-expected-regret}
The best achievable worst-case expected regret as a function of the concept class $\mathcal{C}$ and the time horizon $\operatorname{T}$, denoted by $\mathbf{R}^{\star}(\mathcal{C}, \operatorname{T})$, is defined as follows:
\begin{equation*}
    \mathbf{R}^{\star}(\mathcal{C}, \operatorname{T}) := \inf_{\mathbf{A} \, \in \, \mathcal{A}} \;\; \sup_{\mathcal{S} \, \in \, (\mathcal{X} \times \mathcal{Y})^{\operatorname{T}}} \; \mathbf{R}(\mathbf{A}; \mathcal{S}; \mathcal{C}),
\end{equation*}
where $\mathcal{A}$ is the set of all randomized adversary-directed online learning rules and $\mathcal{S}$ is deterministically labeled.
\end{definition}

\paragraph{Oblivious vs.\ Adaptive Adversaries.} Although the above definition is stated for an oblivious adversary, all of our results extend to adaptive adversaries. In particular, in the agnostic setting, we rely on Lemma~4.1 of \cite{cesa2006prediction}.

\subsection{Combinatorial Complexity Parameters} \label{sec: online-parameters}

In this subsection, we define a combinatorial complexity parameter that appears in our online learning results.

\begin{definition}[Finite Rooted Perfect Binary Tree] \label{def: finite_rooted_perfect_binary_tree}
A finite rooted perfect binary tree is a rooted tree with a finite number of nodes in which each non-leaf node has exactly two children and all leaves have the same depth.
\end{definition}

\begin{definition}[Littlestone Tree] \label{def: littlestone-tree}
Let $\mathcal{C} \subseteq \{0, 1\}^{\mathcal{X}}$ be a concept class. A Littlestone tree for $\mathcal{C}$ is a finite rooted perfect binary tree in which each non-leaf node is labeled by an instance from $\mathcal{X}$, and the two outgoing edges of each non-leaf node are labeled by $0$ and $1$, accordingly.
\end{definition}

\begin{definition}[Shattered Littlestone Tree] \label{def: shattered-littlestone-tree}
Let $\mathcal{C} \subseteq \{0, 1\}^{\mathcal{X}}$ be a concept class, and let $\mathcal{T}$ be a Littlestone tree for $\mathcal{C}$. We say that $\mathcal{T}$ is \emph{shattered} by $\mathcal{C}$ if for every root-to-leaf path in $\mathcal{T}$, identified by $\big( (x_1, y_1), (x_2, y_2), \ldots, (x_{\operatorname{T}}, y_{\operatorname{T}}) \big)$ for some $\operatorname{T} \in \mathbb{N} \cup \{0\}$, there exists a concept $c \in \mathcal{C}$ such that $c(x_i) = y_i$ for every $i \in [\operatorname{T}]$.
\end{definition}

\begin{definition}[Littlestone dimension, following \cite{littlestone1988learning, ben2009agnostic}] \label{def: ld}
Let $\mathcal{C} \subseteq \{0, 1\}^{\mathcal{X}}$ be a concept class. The Littlestone dimension of $\mathcal{C}$, denoted by $\LD(\mathcal{C}) \in \overline{\mathbb{N}} \cup \{0, +\infty\}$, is defined by the $\sup_{d \in \mathbb{N} \cup \{0\}}$ such that, there exists a Littlestone tree of depth $d$ for $\mathcal{C}$ that is shattered by $\mathcal{C}$. Also, if $\mathcal{C} = \emptyset$, we have: $\LD(\mathcal{C}) = 0$.
\end{definition}

\section{Online Learning Results} \label{sec: online-results}

In this section, we study optimal expected regret with deterministic labels for classes with finite diameter. First we show that the optimal expected regret with deterministic labels is always upper bounded by the diameter. 

\begin{theorem}
    Let $K \in \mathbb N$. Any concept class $\cC$ such that $D(\cC) \leq K$ satisfies, for every $T \in \Nat$,
    \[
        \bR^\star(\cC,T) \leq K.
    \]
\end{theorem}

\begin{proof}
    Fix any concept $h_0 \in \cC$. We consider a learner $\bA$ that predicts according to $h_0$ on previously unseen instances and memorizes the labels of previously observed instances. Formally, for any $T \in \Nat$, any history $S \in \bigcup_{t=0}^{T-1}(\cX \times \cY)^t$, and any $x \in \cX$, define
    \[
    \bA(T,S,x)
    =
    \begin{cases}
        y & \text{if } (x,y) \in S,\\
        h_0(x) & \text{o.w.}
    \end{cases}
    \]

    Now fix any comparator $h^\star \in \cC$ and any deterministic labeling function $f^\star$. Suppose that on some round the comparator is correct, but $\bA$ is not; that is, $h^\star(x)=f^\star(x)$ while $\bA$ predicts incorrectly on $x$. Since $\bA$ memorizes labels, this can only happen the first time $x$ appears. Moreover, on such a first occurrence, $\bA$ predicts $h_0(x)$, so we must have $h_0(x) \neq h^\star(x)$. Therefore, every such excess-mistake round can be charged to a distinct point in $h_0 \Delta h^\star$. Hence the regret with respect to $h^\star$ is at most
    \[
        |h_0 \Delta h^\star| \leq D(\cC) \leq K.
    \]
    Since this holds for every $h^\star \in \cC$, every deterministic labeling function, and every sequence of instances, we conclude that
    \[
        \bR^\star(\cC,T) \leq K.
    \]
    This completes the proof.
\end{proof}

We then show that, for every $K$, there exists a concept class with Littlestone dimension $1$ and diameter $2K$ such that the optimal expected regret under deterministic labels is $\Omega(K)$ for sufficiently large $T$. This shows that, in contrast to our batch learning results, the dependence of regret of Littlestone classes on $K$ cannot be improved. Our proof relies on the anti-concentration bound for Binomial distributions stated in
\autoref{lem:bin-tail}.

\begin{lemma}[Lemma 4.7.2 of \cite{ash1990information}] \label{lem:bin-tail}
    Let $X$ be a Binomial random variable with parameters $K \in \Nat$ and $p \in (0, 1)$, for any integer $t \geq Kp$ we have
    \[
    \prob{X \geq t} \geq \sqrt{\frac{K}{8t(K - t)}} \exp \inparen{-K \KL \inparen{\frac{t}{K} \Big \| p }}
    \]
    where $\KL \inparen{q \| p } := q \ln \frac{q}{p} + (1 - q) \ln \frac{1 - q}{p}$ is the KL divergence of Bernoulli random variables with parameters $q$ and $p$.
    
\end{lemma}

\begin{theorem}
    For every $K \in \Nat$, there exists a concept class $\cC$ such that $\LD(\cC)=1$ and $D(\cC)=2K$, and for every $T \ge K$,
\[
    \bR^\star(\cC,T)
    \ge
    \floor{
        \frac{1}{4}
        \min\!\inbrace{
            2\sqrt{K \ln\!\inparen{\frac{T}{4K\sqrt K}}},
            K
        }
    }.
\]
In particular, for $T \ge 4K\sqrt K \exp\!\inparen{\sqrt K}$, we have
\[
    \bR^\star(\cC,T) \ge \floor{\frac{K}{4}}.
\]
\end{theorem}
\begin{proof}
    Consider the concept class $\cC = \inbrace{h_i \mid i \in \Nat \cup \{0\}}$ over the natural numbers, where for every $x \in \Nat$,
    \[
        h_i(x) := \mathbbm{1}\inbrace{x \in [(i - 1) K + 1, K i] }
        \quad \text{for } i \in \Nat,
        \qquad
        h_0(x) := 0 .
    \]
    Clearly, $\LD(\cC) = 1$ and $D(\cC) = 2K$.
    Choose
    \(
        \cS = \inbrace{(1,y_1),(2,y_2),\ldots,(T,y_T)},
    \)
    where $y_t$ is drawn i.i.d. from a Bernoulli distribution with parameter $\frac{1}{2}$.
    Note that for every $t \in [T]$ and every $y \in \{0,1\}$, we have
    $\prob{y \neq y_t} = \nicefrac{1}{2}$.
    Therefore, every learner $\bA$ satisfies
    \begin{equation}
    \label{eq:online-lower-mistake}
        \cexpect{\cS}{\widebar{\bM}(\cA;\cS)} = \frac{T}{2}.
    \end{equation}

    We then attempt to derive a lower bound for
    \[
     \inf_{i \in \Nat \cup 0} \, \sum_{t = 1}^{T} \, \mathds{1} \insquare{ h_i(x_t) \neq y_t  }.
    \]
    Denote $m := \floor{\frac{T}{K}}$, and for every $i \in [m]$, denote
    \(
        Z_i = \sum_{t = (i - 1)K + 1}^{iK} y_t .
    \)
    Then observe that the all-zero function $h_0$ makes $\sum_{t = 1}^T y_t$ mistakes.
    For every $i \in [m]$, the function $h_i$ makes the same predictions as $h_0$, except on
    $[(i - 1)K + 1, Ki]$.
    On this block, $h_0$ makes $Z_i$ mistakes, while $h_i$ makes $K - Z_i$ mistakes.
    Therefore, in total, $h_i$ makes
    \[
        \sum_{t = 1}^T y_t - Z_i + (K - Z_i)
        =
        \sum_{t = 1}^T y_t + K - 2Z_i
    \]
    mistakes. This implies that
    \[
    \begin{aligned}
        \inf_{i \in \Nat \cup 0} \, \sum_{t = 1}^{T} \, \mathds{1} \insquare{ h_i(x_t) \neq y_t  } & \leq \min_{i \in \inbrace{0, 1, \cdots, m}} \, \sum_{t = 1}^{T} \, \mathds{1} \insquare{ h_i(x_t) \neq y_t }  \\
        % & =
        % \min \inbrace{\min_{i \in [m]} \inparen{ \sum_{t \notin \insquare{(i - 1)K + 1, iK}} y_t + \sum_{t \in \insquare{(i - 1)K + 1, iK}} \inparen{1 - y_t}}, } \\
        % & = \inf_{i \in [m]} \sum_{t = 1}^{m} y_t + \sum_{t \in \insquare{(i - 1)K + 1, iK}} 1 - 2 y_t \\
        % & =
        % \min \inbrace{\min_{i \in [m]} \inparen{ \sum_{t = 1}^m y_t + K - 2 Z_i}, \sum_{t = 1}^m y_t} \\
         & = \sum_{t = 1}^T y_t - \max(2 \max_{i \in [m]} Z_i - K, 0)
    \end{aligned}
    \]
    Thus,
    \[
    \cexpect{\cS}{\inf_{i \in \Nat} \, \sum_{t = 1}^{T} \, \mathds{1} \insquare{ h_i(x_t) \neq y_t  }} \leq \frac{T}{2} - \cexpect{\cS}{\max(2 \max_{i \in [m]} Z_i - K, 0)}
    \]
    Combining with \autoref{eq:online-lower-mistake} we get
    \begin{equation} \label{eq:online-lower-regret}
        \cexpect{\cS}{\widebar{\bM}(\cA; \cS) - \inf_{i \in \Nat} \, \sum_{t = 1}^{T} \, \mathds{1} \insquare{ h_i(x_t) \neq y_t  }} \geq \cexpect{\cS}{\max(2 \max_{i \in [m]} Z_i - K, 0)}
    \end{equation}

    We now lower bound the right-hand side. Applying \autoref{lem:bin-tail}, and for every $i \in [m]$ and every integer $r$ such that $0 \leq r \leq \frac{K}{4}$, we derive
    \[
    \begin{aligned}
        p_r := \prob{Z_i \geq \frac{K}{2} + r} &\geq \sqrt{\frac{K}{8 r (K - r)}} \exp \inparen{-K \KL \inparen{\frac{1}{2} + \frac{r}{K} \Big \| \frac{1}{2}}} \\
        & \geq \frac{1}{2\sqrt{K}} \exp \inparen{-K \KL \inparen{\frac{1}{2} + \frac{r}{K} \Big \| \frac{1}{2}}}
    \end{aligned}
    \]
    Using the elementary inequality  $\KL \inparen{\frac{1}{2} + x \Big \| \frac{1}{2}} \leq 4x^2$ for $x \leq \frac{1}{4}$, we get
    \[
    \prob{Z_i \geq \frac{K}{2} + r} \geq \frac{1}{2\sqrt{K}} \exp \inparen{-\frac{4r^2}{K}}.
    \]
    Set $r := \floor{\frac{1}{4} \min\inbrace{ 2 \sqrt{K \ln \inparen{\frac{m}{2 \sqrt{K}}}}, K}}$. Then,
    \[
    \frac{4 r^2}{K} \leq \ln \inparen{\frac{m}{2\sqrt{K}}}
    \]
    This implies $p_r \geq m^{-1}$. Therefore,
    \[
    \prob{\max_{i \in [m]} Z_i \geq \frac{K}{2} + r} = 1 - (1 - p_r)^{m} \geq 1 - e^{-m p_r} \geq 1 - e^{-1} \geq \frac{1}{2}.
    \]
    This indicates that
    \[
    \expect{\max(2\max_{i \in [m]} Z_i - K, 0)} \geq \frac{1}{2}.2r + \frac{1}{2}.0 = r
    \]
    % Observe that since $T \geq K$, we have $m \geq \frac{T}{2K}$. Thus, 
    Plugging this into \autoref{eq:online-lower-regret}, for every learner $\bA$ we get
    \[
      \cexpect{\cS}{\widebar{\bM}(\cA; \cS) - \inf_{i \in \Nat} \, \sum_{t = 1}^{T} \, \mathds{1} \insquare{ h_i(x_t) \neq y_t  }} \geq r \geq \floor{\frac{1}{4} \min\inbrace{ 2 \sqrt{K \ln \inparen{\frac{T}{4 K \sqrt{K}}}}, K}}
    \]
    Where the second in equality is due the fact that for $T \geq K$, we have $m \geq \frac{T}{2K}$. This implies that, for every learner $\bA$ there exists a sequence $\cS$ such that 
    \[
    \bR(\bA; \cS, \cC) \geq \floor{\frac{1}{4} \min\inbrace{ 2 \sqrt{K \ln \inparen{\frac{T}{4 K \sqrt{K}}}}, K}},
    \]
    which completes the proof.
    % Plugging this into \autoref{eq:online-lower-regret} completes the proof.
\end{proof}

\section{Useful Theorems}

\begin{lemma}[Multiplicative Chernoff bounds \cite{motwani1996randomized}] \label{lem:mult-chern-bound} Let $X_1, \ldots, X_m$ be independent random variables drawn according to some distribution $\mathcal{D}$ with mean $p$ and support included in $[0,1]$. Then, for any $\gamma \in\left[0, \frac{1}{p}-1\right]$, the following inequality holds for $\widehat{p}=\frac{1}{m} \sum_{i=1}^m X_i$:

\[
\begin{aligned}
& \mathbb{P}[\widehat{p} \geq(1+\gamma) p] \leq e^{-\frac{m p \gamma^2}{3}} \\
& \mathbb{P}[\widehat{p} \leq(1-\gamma) p] \leq e^{-\frac{m p \gamma^2}{2}}
\end{aligned}
\]
\end{lemma}

\begin{theorem}[Hoeffding Inequality \cite{hoeffding1994probability}] \label{thm:Hoeffding}
    Let $X_1, \ldots, X_n$ be independent random variables such that $a_i \leq X_i \leq b_i$ almost surely. Consider the sum of variables,
\[
S_n=X_1+\cdots+X_n
\]

Then Hoeffding's theorem states that, for all $t>0$,

\[
\begin{gathered}
\mathrm{Pr}\left(\left|S_n-\mathrm{E}\left[S_n\right]\right| \geq t\right) \leq 2 \exp \left(-\frac{2 t^2}{\sum_{i=1}^n\left(b_i-a_i\right)^2}\right)
\end{gathered}
\]

\end{theorem}

\begin{lemma}[Sedrakyan inequality \cite{sedrakyan1997applications, sedrakyan2018algebraic}] Let $a_1, a_2, \ldots, a_n$ be real numbers and $b_1, b_2, \ldots, b_n$ positive real numbers. Then
\[
\frac{a_1^2}{b_1}+\cdots+\frac{a_n^2}{b_n} \geq \frac{\left(a_1+\cdots+a_n\right)^2}{b_1+\cdots+b_n}
\]   
\end{lemma}
\clearpage

% \input{checklist}
% \clearpage

\end{document}